\documentclass[11pt]{article}

\usepackage[preprint]{acl}

\usepackage{times}
\usepackage{latexsym}

\usepackage[T1]{fontenc}

\usepackage[utf8]{inputenc}

\usepackage{microtype}

\usepackage{inconsolata}

\usepackage{graphicx}

\usepackage{booktabs}
\usepackage{url}
\usepackage{times}
\usepackage{latexsym}

\usepackage{amsmath,amssymb}
\usepackage{graphicx}
\usepackage{tikz}
\usetikzlibrary{positioning,arrows.meta,fit,shapes.geometric}
\usepackage{adjustbox}
\usepackage{tabularx}
\usepackage{ragged2e}
\usepackage{multirow}
\usepackage{pifont}
\usepackage{soul}

\usepackage[toc,page]{appendix}
\usepackage{xcolor}
\newcommand{\arbor}{\textsc{Arbor}}

\usepackage{color}
\usepackage{floatrow}
\usepackage{makecell}
\usepackage{xspace}
\usepackage{colortbl}
\usepackage{enumitem}
\usepackage[most]{tcolorbox}

\definecolor{airforceblue}{rgb}{0.36, 0.54, 0.66}
\definecolor{ao(english)}{rgb}{0.0, 0.5, 0.0}
\definecolor{auburn}{rgb}{0.43, 0.21, 0.1}
\definecolor{bittersweet}{rgb}{1.0, 0.44, 0.37}

\usepackage[tableposition=top]{caption}
\usepackage{floatrow}
\usepackage{subcaption}
\usepackage{wrapfig}
\usepackage{makecell}

\newfloatcommand{capbtabbox}{table}[][\FBwidth]

\usepackage{arydshln}

\usepackage{minibox}
\usepackage{tabularx,ragged2e}
\usepackage{tabularx} %
\usepackage{algorithm}
\usepackage[noend]{algpseudocode}
\usepackage{bbm}

\definecolor{casehl}{HTML}{FDE2CF}
\newcommand{\cl}[3]{{\color{gray}\tiny #1} & #2\hspace*{\fill}{\color{gray}\itshape\ #3} \\}
\newcommand{\ev}[3]{\rowcolor{casehl}{\color{gray}\tiny #1} & \textbf{#2}\hspace*{\fill}{\color{gray}\itshape\ #3} \\}
\newcommand{\el}{\multicolumn{2}{@{}l}{{\color{gray}\tiny \quad$\cdots$}} \\}
\newcommand{\caseblock}[4][0.49\linewidth]{\begin{minipage}[t]{#1}\scriptsize\setlength{\tabcolsep}{2pt}\renewcommand{\arraystretch}{1.05}
\textbf{#2}\\ {\color{gray}\itshape Predictive state: #3}\\[2pt]
\begin{tabular}{@{}r>{\raggedright\arraybackslash}p{0.94\linewidth}@{}}
#4
\end{tabular}\end{minipage}}

\usepackage{placeins}

\usepackage{amsmath,amsfonts,bm}

\def\eqref#1{equation~\ref{#1}}

\def\1{\bm{1}}

\DeclareMathAlphabet{\mathsfit}{\encodingdefault}{\sfdefault}{m}{sl}
\SetMathAlphabet{\mathsfit}{bold}{\encodingdefault}{\sfdefault}{bx}{n}

\title{\textsc{Arbor}: Locating Reasoning Failures in Long Chains of Thought through Reasoning State Tracking}

\author{
  \textbf{Huihan Li\textsuperscript{1}},
  \textbf{Siyuan Wang\textsuperscript{2}},
  \textbf{Naga Vamsi Ramana Dinavahi\textsuperscript{1}},
  \textbf{Xiaohui Liu\textsuperscript{3}},
  \textbf{Xiang Ren\textsuperscript{1}}
\\
  \textsuperscript{1}University of Southern California,
  \textsuperscript{2}The Chinese University of Hong Kong,
  \textsuperscript{3}New York University
\\
  \small{
    \texttt{\{huihanl,xiangren\}@usc.edu},
    \texttt{siyuanwang@cuhk.edu.hk}
  }
}

\begin{document}
\maketitle
\begin{abstract}
Large reasoning models solve complex problems through long chains of thought, in which prior work has documented failure modes such as overthinking, underthinking, and unsupported deductions. Existing methods detect these failures from \emph{local} signals: textual similarity to nearby steps, switch tokens, or an LLM's judgment of a step's correctness. However, what a step contributes to the solution depends on the state that earlier steps have established, so identifying failures requires a \emph{global} view that tracks that state across the trajectory. Therefore, we propose \arbor{}, a reasoning-failure attribution framework that represents each trajectory as a \emph{tree of reasoning states}, built as each step creates, verifies, prunes, or revisits a state, and attributes each failure to a relation between a step and the state preceding it. Using \arbor{}, we refine the three failure modes into distinct subtypes that harm reasoning through different mechanisms and at different costs. We find that the tree state \arbor{} maintains before a step separates failure steps from the at-risk steps at which the failure did not occur. We discuss how \arbor{} integrates with existing mitigations at inference time and in training data.
Code, data and interactive examples are available at \url{https://inklab.usc.edu/ARBOR}.

\end{abstract}

\section{Introduction}
Large reasoning models (LRMs) solve complex reasoning tasks with long chains of thought (CoT) in which they derive intermediate results, verify them, backtrack and pivot to alternatives. Prior work has documented recurring failure modes in these trajectories: overthinking, in which the model re-reasons about results it has already established \citep{chen2025overthinking,pipis2026loop}; underthinking, in which it abandons a line of reasoning that could still succeed, a \emph{premature} switch as opposed to a \emph{justified} switch away from a line already ruled out \citep{wang2025underthinking}; and unsupported deductions, steps the preceding context does not entail \citep{lightman2024verify,he2025deltabench}. These failures waste inference budget and some produce wrong answers; reducing them requires knowing which steps fail and \textit{why}.

Existing methods detect these failures with signals that are \textbf{local} to a step: recurring $n$-grams or embedding similarity between nearby sentences for repetition \citep{pipis2026loop,huang2026seer,zhen2025reconsidering}, thought-switch tokens such as \emph{alternatively} for underthinking \citep{wang2025underthinking,zhang2025smartswitch}, process reward models and LLM judges for step correctness \citep{wang2024mathshepherd,zhang2025lessons,zheng2025processbench}, and block-level functional labels spanning many steps \citep{marjanovic2026thoughtology,luo2025dlcot,jiang2025lcot2tree,bogdan2025thoughtanchors}. None conditions a step's label on which earlier steps it depends on and what they established, although that context is what makes a derivation a repetition, a switch premature or a deduction unsupported (related work in Appendix~\ref{sec:related}).

\definecolor{figred}{RGB}{220,38,38}
\definecolor{figorange}{RGB}{233,113,50}
\definecolor{figpurple}{RGB}{124,58,237}
\definecolor{figgreen}{RGB}{21,128,61}
\begin{figure*}[t]
    \centering
    \includegraphics[width=\linewidth]{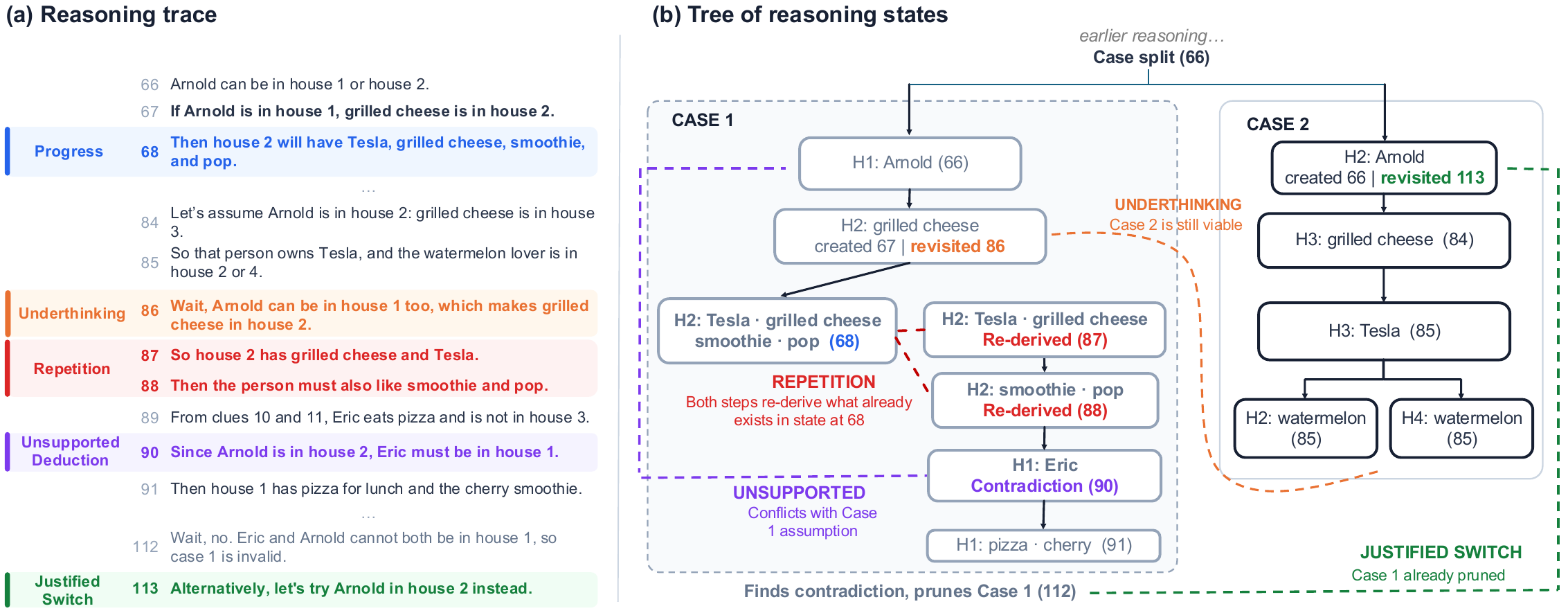}
    \caption{\textbf{How a step relates to the states already established defines a reasoning failure.} (a)~An excerpt of a \texttt{Qwen3-8B} trajectory on a ZebraLogic puzzle; numbers are step indices, and colored steps are labeled by function. (b)~Its tree of reasoning states: nodes are intermediate states, tagged with the steps that created or revisited them, edges link each state to the states derived from it, and the case split at step~66 opens two branches. Steps~87--88 are a \textcolor{figred}{\textbf{repetition}}: they re-derive the state of step~68 in new words. Step~86 is \textcolor{figorange}{\textbf{underthinking}}: it abandons Case~2 while that branch is still viable. Step~90 is an \textcolor{figpurple}{\textbf{unsupported deduction}}: it contradicts the Case~1 assumption. Step~113 is a \textcolor{figgreen}{\textbf{justified switch}}: it follows the pruning of Case~1 at step~112.}
    \label{fig:teaser}
\end{figure*}

\begin{figure*}[t]
    \centering
    \includegraphics[width=\linewidth]{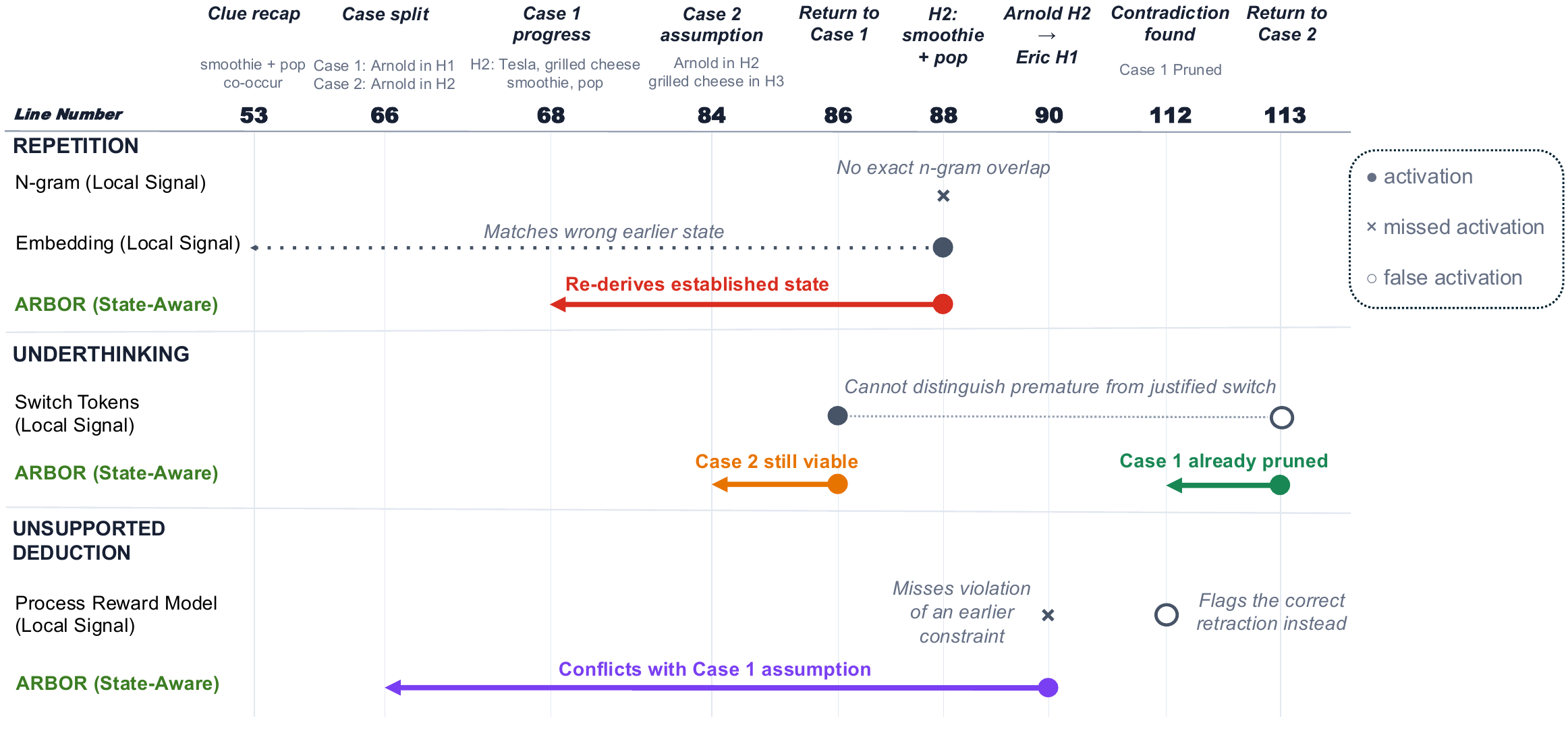}
    \caption{\textbf{Local signals miss or misattribute the failures that the reasoning state identifies.} Columns are steps of the trajectory in Figure~\ref{fig:teaser}. For each failure mode, local signals (gray, dotted) are compared with \arbor{} (colored, solid), whose arrow points to the earlier step whose state determines the label. \emph{Repetition:} step~88 shares no $n$-gram with step~68, and embedding similarity links it to the clue recap at step~53 rather than to the state it re-derives. \emph{Underthinking:} a switch token fires at both step~86 and step~113, but only step~86 leaves a viable branch. \emph{Unsupported deduction:} the process reward model misses step~90, which violates the Case~1 assumption of step~66, and flags the correct retraction at step~112 instead.}
    \label{fig:comparison}
\end{figure*}

We therefore define reasoning failures relative to the \textit{state of the reasoning}, the intermediate results established, the branches open and the constraints in force when a step occurs. We represent each trajectory as a tree of reasoning states (Figure~\ref{fig:teaser}), in which nodes are intermediate results, each edge records the node a step derives from, and where a step attaches identifies a failure; Figure~\ref{fig:comparison} shows why local signals fail on the same trajectory.

We make these relations explicit with \arbor{} (\textbf{A}ttributing \textbf{R}easoning failures \textbf{B}y \textbf{O}perations on \textbf{R}easoning-state trees), a pipeline in three stages (Figure~\ref{fig:arbor}; \S\ref{sec:methodology}): an LLM \textit{parses} each step into an operation on a canonical, wording-invariant payload; a deterministic \textit{engine} resolves the step against the tree built so far and applies the corresponding mutation; and we \textit{define failures} as relations between a step and a specific earlier state, which refines the three failure modes into distinct subtypes (Table~\ref{tab:error-rows}).

We instantiate the framework on three reasoning tasks with distinct reasoning structures, constraint satisfaction (ZebraLogic; \citealp{lin2025zebralogic}), search over arithmetic compositions (24~Game\footnote{nlile/24-game}) and sequential action planning under preconditions (PlanBench; \citealp{valmeekam2023planbench}), and report three findings.

First, \arbor{} flags failures more accurately than local signals (\S\ref{sec:results-local}). The local signals recover only a small share of the steps \arbor{} flags, and what they flag instead is mostly ordinary reasoning. \arbor{}'s per-subtype counts separate incorrect from correct trajectories better than all local signals combined.

Second, the distinct subtypes of each failure mode differ in cost and mechanism (\S\ref{sec:results-subtypes}). Unsupported deduction costs correctness, since the model's verification seldom catches a step the state does not support, and the step enters the answer. Underthinking costs the search expended before the model reaches the answer by another path or by returning to the branch, rather than correctness, since it usually still reaches it. Repetition does the least harm to the answer and costs few tokens beyond its own steps, since the model re-establishes a step it proposed long before and gains nothing from it. Within each mode, the subtypes differ in which of these costs they carry.

Third, the tree state before a step separates each modeled subtype from the at-risk steps at which it did not end up occurring (\S\ref{sec:results-precursors}). The precursor of each modeled subtype is the content of the branch, whether the current node has been verified, or the recent operations. Operation labels alone recover only the last kind, and the first two precede the subtypes that cost correctness or viable work, which therefore require the tree.

To our knowledge, \arbor{} is the first reasoning-failure attribution framework that maintains the solution state of a reasoning trajectory step by step and attributes each failure to a relation between a step and that state, rather than to the step's text. Tracking the state enables us to separate the subtypes of a failure mode, measure what each costs in tokens and in correctness, and identify the state that precedes a failure. We discuss how \arbor{} integrates with existing mitigation at inference time and in training data (\S\ref{sec:results-precursors}).

\section{Methodology}
\label{sec:methodology}

\begin{figure}[t]
    \centering
    \includegraphics[width=0.5\textwidth]{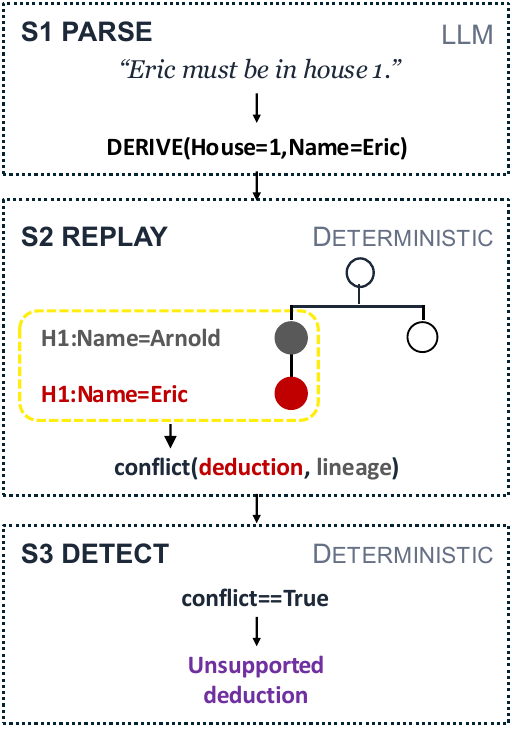}
    \caption{\textbf{\arbor{}'s three-stage framework.} An LLM parses each step into a canonical operation, then a deterministic engine replays it against the accumulated reasoning state, and deterministic rules map the resulting state relations to functional labels.}
    \label{fig:arbor}
\end{figure}
\begin{table*}[t]
\centering
\small
\setlength{\tabcolsep}{5pt}
\begin{tabular}{@{}l p{0.72\textwidth}@{}}
\toprule
\textbf{Row} & \textbf{Definition: the model \ldots} \\
\midrule
\multicolumn{2}{@{}l}{\textit{Repetition}} \\
R1 Re-Derivation & derives again a result it has already derived, after deriving or verifying others. \\
R2 Derive-after-Prune & derives from a hypothesis it has already rejected. \\
R3 Redundant Re-Verification & verifies a result again although nothing it depends on has changed. \\
R4 Subgoal Churn & posts a subgoal again after working on others, without progress on it. \\
\midrule
\multicolumn{2}{@{}l}{\textit{Underthinking}} \\
R5 Branch Abandonment & abandons a hypothesis that can still reach a correct solution, even if it later returns. \\
R6 Subgoal Abandonment & drops a subgoal that is still achievable and never completes it. \\
R7 Post-Answer Attempt & begins searching elsewhere, or re-verifying, while holding a correct complete solution. \\
\midrule
\multicolumn{2}{@{}l}{\textit{Unsupported Deduction}} \\
R8 Premature Finalization & commits an answer that its established results or the task contradict. \\
R9 Infeasible Derivation & derives a result that its established results do not permit. \\
R10 False Rejection & rejects a hypothesis that can still reach a correct solution, for a false reason. \\
R11 False Verdict & records a verification outcome that its established results contradict. \\
\bottomrule
\end{tabular}
\caption{The failure rows. Each row is one relation between a step and the results the model has established before it; \S\ref{sec:errors} gives the predicate on the reasoning-state tree. A row is undefined, rather than approximated, on a task whose structure has no state for it (Appendix~\ref{sec:app-subtypes}).}
\label{tab:error-rows}
\end{table*}

\arbor{} (Attributing Reasoning failures By Operations on Reasoning-state trees) turns a trajectory $x_1,\dots,x_T$ into a tree of reasoning states in three stages (Figure~\ref{fig:arbor}): a \emph{parser} maps each step to an operation on a schema-grounded payload; a deterministic \emph{engine} resolves the step to a node of the tree built so far and applies one mutation; and a \emph{failure row} (Table~\ref{tab:error-rows}) is a predicate on the step, the node it resolves to and the tree before it.
Only the parser calls a model; the same parse always replays to the same tree and the same labels, and every label names the earlier node it is defined against.

\subsection{Parser}
\label{sec:parser}

\begin{table}[t]
\centering
\footnotesize
\setlength{\tabcolsep}{3pt}
\begin{tabular}{@{}>{\raggedright\arraybackslash}p{0.20\linewidth}>{\raggedright\arraybackslash}p{0.31\linewidth}>{\raggedright\arraybackslash}p{0.40\linewidth}@{}}
\toprule
\textbf{Operation} & \textbf{Payload} & \textbf{Effect on the tree} \\
\midrule
\textsc{Derive} & a result $f$, or alternatives $\{f_i\}$ & adds a child holding $f$ under the node $f$ depends on; one child per $f_i$ \\
\textsc{Assume} & a hypothesis $h$ & adds a child holding $h$ under the current leaf \\
\textsc{Verify} & a result $f$, verdict $o$ & records $o$ and $s(u)\models f$ on the node $u$ holding $f$ \\
\textsc{Prune} & a result or hypothesis $f$ & closes the node holding $f$ and its subtree \\
\textsc{Confirm} & an answer $a$ & commits the node holding $a$; records $\mathrm{accept}(a)$ \\
\textsc{Restate} & a result $f$ & records the restatement on the node holding $f$ \\
\textsc{Meta} & none & none \\
\bottomrule
\end{tabular}
\caption{The seven operations: the parser assigns each step one operation and its payload, and the engine applies the effect. A result is a fact, value, action or subgoal.}
\label{tab:operations}
\end{table}

The first stage, the parser, assigns every step one of seven \emph{operations} (Table~\ref{tab:operations}) together with its \emph{payload}, the schema-grounded content the operation acts on: a fact, value, action, subgoal, hypothesis or answer.
\textsc{Derive} adds a result, \textsc{Assume} opens a branch, \textsc{Verify} records a verdict on a result, \textsc{Prune} closes a branch, \textsc{Confirm} commits an answer, \textsc{Restate} repeats a result without deriving it, and \textsc{Meta} carries no content.
For each task we customize the parser's prompt with a \emph{puzzle schema}, the entities, the allowed predicates, the allowed operations with their preconditions and effects, the initial state and the goal, and the parser outputs in that schema, so ``Bob lives in the fifth house'' and ``house 5's name is Bob'' parse to the same atom.
We use Qwen3.8-27B as the parser; it sees the puzzle schema, the current step, its labeled neighborhood and the schema elements of the previous steps.
The parser reaches an accuracy of 0.89 or above on adjudicated reference labels on every task (Appendix~\ref{sec:appendix_parser}).
The parser thus records only what each step does, in isolation.

\subsection{Engine}
\label{sec:engine}

From these per-step operations the engine builds the tree of reasoning states, applying each step to the tree built from the steps before it.

\paragraph{Tree.}
A \emph{node} $v$ holds an intermediate result or hypothesis (a cell assignment, a value composed from a subset of the tiles, a world state after an action) together with the \emph{solution state} $s(v)$ reached at it: the board with every cell placed so far, the tiles still unused, the facts that hold after the actions taken.
The \emph{lineage} of $v$ is the path from the root to $v$.
A \emph{branch} is the subtree below a hypothesis.
A node's \emph{status} is open, closed by the model, contradictory as detected by the engine, or committed.
The \emph{frontier} is the set of open leaves, and the \emph{current leaf} is the frontier leaf at which the model is working.
The \emph{dependency path} of an answer is the lineage of its committed node.

\paragraph{Applying a step.}
On this tree, the engine applies a step by resolving the node its payload refers to, $r(x_t)\in\mathcal{T}_{t-1}$, and applies the operation's effect (Table~\ref{tab:operations}).
A \textsc{Derive} resolves to the node its content depends on, or to the current leaf when it names none; every other operation resolves to the node holding its payload (procedure in Appendix~\ref{sec:appendix_resolution}).
A \textsc{Verify} or \textsc{Restate} records the claim beside what the node holds, so that a claim the state contradicts stays visible to the detectors.

\paragraph{Oracles.}
On the solution state of each node, the engine also evaluates three \emph{oracles}, task-specific predicates computed from that state and the task definition: $\mathrm{sound}(v,f)$, whether a derived fact $f$ follows from $s(v)$ and the problem's rules; $\mathrm{viable}(v)$, whether a correct complete solution is still reachable from $v$ (the puzzle's unique solution; the arithmetic over the remaining tiles; a planner run from $s(v)$), a node that is not viable being \emph{dead}; and $\mathrm{accept}(a)$, whether a committed answer $a$ is correct and complete.

\subsection{Failure Detectors}
\label{sec:errors}

With the tree and its oracles in place, the detectors read each step against them.
Each row of Table~\ref{tab:error-rows} is a predicate $R(x_t,\,v,\,\mathcal{T}_{t-1})$ on the step, its resolved node $v=r(x_t)$ and the tree before it, and a step on which the predicate holds is an \emph{event} of that row.
In the tree's terms, a hypothesis of Table~\ref{tab:error-rows} is a branch, the established results are the solution state $s(v)$, and a solution is reachable when $\mathrm{viable}(v)$ holds.
The three failure modes differ in what the predicate tests.
Repetition rows test whether a node in $\mathcal{T}_{t-1}$ already holds what the step produces, and which such node: one worked away from, one closed, one already verified, or a subgoal already posted.
Underthinking rows test $\mathrm{viable}(v)$ on the node the step leaves, so a switch away from a dead node is justified and fires no row, or whether $v$ already holds an accepted solution (the literature's overthinking, filed here because the step leaves a finished solution).
Unsupported-deduction rows test the step against the oracles: $\mathrm{sound}$ on a derivation, $\mathrm{viable}$ on a pruned branch, $\mathrm{accept}$ on a committed answer, and the recorded verdict of a verification against $s(v)$.
The validation of the engine and the detectors is in Appendix~\ref{sec:appendix_validation}.

\section{Experimental Setup}
\label{sec:results-setup}

\paragraph{Tasks.}
We use three reasoning tasks with distinct reasoning structures: ZebraLogic \citep{lin2025zebralogic}, constraint satisfaction; the 24~Game\footnote{nlile/24-game}, search over arithmetic compositions; and PlanBench \citep{valmeekam2023planbench}, sequential action planning under preconditions (Details in Appendix~\ref{sec:setup}).

\paragraph{Models and sampling.}
We study the Qwen3 family; all results are on trajectories of \texttt{Qwen3-8B}, and the comparison across model sizes is in Appendix~\ref{sec:app-scale}.
We sample each question three times, with a budget of 8{,}192 generated tokens on ZebraLogic and the 24~Game and 16{,}000 on PlanBench.

\paragraph{Evaluation.}
A trajectory ends correct, incorrect or truncated at the token budget; the scoring rule per task and the outcome shares per task and model size are in Appendix~\ref{sec:setup}, Table~\ref{tab:main_results}.

\begin{figure}[t]
\centering
\small
\resizebox{\linewidth}{!}{%
\begin{tikzpicture}[
  x=1cm, y=0.55cm,
  n/.style={circle, draw=black!75, fill=white, inner sep=0pt, minimum size=5.5pt},
  p/.style={circle, draw=black!30, dashed, fill=white, inner sep=0pt, minimum size=5.5pt},
  bad/.style={circle, draw=red!75, very thick, fill=red!25, inner sep=0pt, minimum size=5.5pt},
  ans/.style={circle, draw=black!80, fill=black!80, inner sep=0pt, minimum size=5.5pt},
  e/.style={draw=black!65, thick},
  pe/.style={draw=black!25, dashed},
  re/.style={-{Stealth[length=3pt]}, draw=red!75, thick, dashed, bend right=45},
  lab/.style={font=\footnotesize, text=black!80, align=center},
  fmark/.style={font=\footnotesize, text=red!75, align=left},
  note/.style={font=\footnotesize, text=red!75, align=left}
]
\begin{scope}[xshift=0cm]
  \node[lab] at (0.6,0.9) {ZebraLogic};
  \node[n] (z1) at (0.6,0) {};
  \node[n] (z2) at (0.6,-1) {};
  \node[n] (z3) at (0.6,-2) {};
  \node[p] (z4b) at (1.35,-3) {};
  \node[n] (z4) at (0.3,-3) {};
  \node[n] (z5) at (0.3,-4) {};
  \node[n] (z6) at (0.3,-5) {};
  \node[p] (z7b) at (1.05,-6) {};
  \node[n] (z7) at (0.0,-6) {};
  \node[n] (z8) at (0.0,-7) {};
  \node[n] (z9) at (0.0,-8) {};
  \draw[e] (z1)--(z2)--(z3)--(z4)--(z5)--(z6)--(z7)--(z8)--(z9);
  \draw[pe] (z3)--(z4b); \draw[pe] (z6)--(z7b);
  \node[lab, anchor=west, text=black!60] at (0.85,-1.95) {case split};
  \draw[re] (z8) to (z4);
  \node[note, anchor=west] at (0.75,-7.3) {re-derives};
  \node[lab, anchor=north west, align=left] at (-0.3,-9.0) {Shape: deep chain,\\\hphantom{Shape: }case splits\\Error: out of\\\hphantom{Error: }budget};
\end{scope}
\begin{scope}[xshift=3.0cm]
  \node[lab] at (0.9,0.9) {24 Game};
  \node[n] (g0) at (0.9,0) {};
  \node[p] (g1) at (-0.1,-1.6) {}; \node[p] (g2) at (0.3,-1.6) {}; \node[n] (g3) at (0.7,-1.6) {};
  \node[p] (g4) at (1.1,-1.6) {}; \node[n] (g5) at (1.5,-1.6) {}; \node[p] (g6) at (1.9,-1.6) {};
  \draw[pe] (g0)--(g1); \draw[pe] (g0)--(g2); \draw[e] (g0)--(g3); \draw[pe] (g0)--(g4); \draw[e] (g0)--(g5); \draw[pe] (g0)--(g6);
  \node[n] (g32) at (0.7,-3.2) {};
  \node[bad] (g51) at (1.5,-3.2) {};
  \node[ans] (g52) at (1.5,-4.8) {};
  \draw[e] (g3)--(g32); \draw[e] (g5)--(g51)--(g52);
  \node[lab, anchor=north east, align=right, text=black!60] at (0.95,-3.45) {correct,\\not committed};
  \node[fmark, anchor=west] at (1.75,-3.2) {infeasible};
  \node[lab, anchor=west, text=black!60] at (1.75,-4.8) {committed};
  \node[lab, anchor=north west, align=left] at (-0.3,-9.0) {Shape: shallow fan,\\\hphantom{Shape: }most pruned\\Error: wrong one\\\hphantom{Error: }of two};
\end{scope}
\begin{scope}[xshift=6.0cm]
  \node[lab] at (0.7,0.9) {PlanBench};
  \node[n] (p1) at (0.7,0) {};
  \node[n] (p2) at (0.7,-1.2) {};
  \node[n] (p3) at (0.7,-2.4) {};
  \node[bad] (p4) at (0.7,-3.6) {};
  \node[n] (p5) at (0.7,-4.8) {};
  \node[n] (p6) at (0.7,-6.0) {};
  \node[ans] (p7) at (0.7,-7.2) {};
  \draw[e] (p1)--(p2)--(p3)--(p4)--(p5)--(p6)--(p7);
  \node[fmark, anchor=west] at (0.95,-3.6) {infeasible step,\\check passes};
  \node[lab, anchor=west, text=black!60] at (0.95,-7.2) {committed};
  \node[lab, anchor=north west, align=left] at (-0.3,-9.0) {Shape: chain of\\\hphantom{Shape: }plan steps\\Error: wrong plan};
\end{scope}
\begin{scope}[shift={(0,-12.9)}]
  \node[n] (l1) at (0.1,0) {}; \node[lab, anchor=west] at (0.25,0) {established state};
  \draw[pe] (4.0,0)--(4.5,0); \node[lab, anchor=west] at (4.6,0) {pruned branch};
  \node[ans] (l3) at (0.1,-1.3) {}; \node[lab, anchor=west] at (0.25,-1.3) {committed answer};
  \node[bad] (l4) at (4.25,-1.3) {}; \node[lab, anchor=west] at (4.6,-1.3) {Infeasible Derivation};
\end{scope}
\end{tikzpicture}}
\caption{\textbf{Three tasks, three shapes of reasoning, three failure patterns.} Schematic \arbor{} trees of a typical failing trajectory for each task (statistics in Appendix~\ref{sec:app-tree-shapes}).}
\label{fig:tree-shapes}
\end{figure}
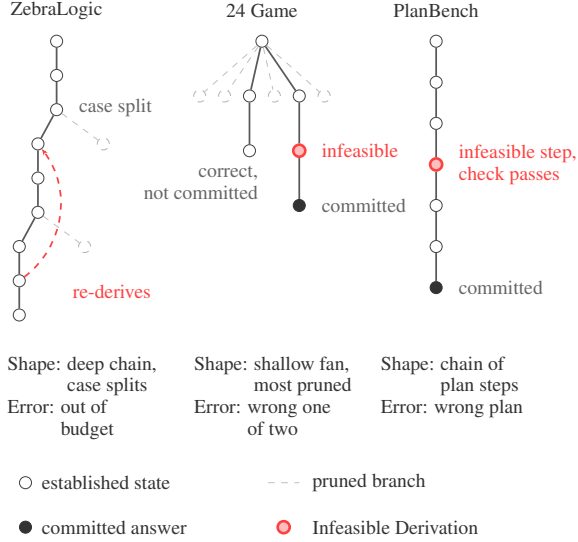

\paragraph{Tree shapes and failure patterns.}
Read off \arbor{}'s trees, a ZebraLogic trajectory is one deep chain of derivations split at a few case assumptions, a 24~Game trajectory is a shallow fan of alternative compositions tried in turn from the same start, and a PlanBench trajectory is a chain of plan steps that rarely branches (Figure~\ref{fig:tree-shapes}).
Each shape fails in its own way: the ZebraLogic chain is filled with \textbf{Re-Derivation} across its case splits and fails by running out of budget; the 24~Game fan is filled with \textbf{Redundant Re-Verification} and \textbf{Derive-after-Prune} and fails by committing an answer built on an \textbf{Infeasible Derivation} while a correct one stands; the PlanBench chain accumulates infeasible steps that pass their checks and fails as a wrong plan (Tables~\ref{tab:subtypes} and~\ref{tab:main_results}).

\section{\arbor{} versus Local Signals}
\label{sec:results-local}
\label{sec:results-local-setup}

\begin{table}[t]
\centering
\scriptsize
\setlength{\tabcolsep}{2pt}
\begin{tabular}{@{}llrr@{}}
\toprule
\textbf{Category} & \textbf{Method} & \textbf{Prec.} & \textbf{Rec.} \\
\midrule
\multirow{3}{*}{Repetition}
 & \arbor{} & 64\% & 37\% \\
 & n-gram & 49\% & 31\% \\
 & similarity & 62\% & 45\% \\
\midrule
\multirow{2}{*}{Underthinking}
 & \arbor{} & 32\% & 71\% \\
 & switch tokens & 27\% & 62\% \\
\midrule
\multirow{2}{*}{Post-Answer Attempt}
 & \arbor{} & 77\% & 87\% \\
 & ERD & 11\% & 13\% \\
\midrule
\multirow{3}{*}{\shortstack[l]{Unsupported\\deduction}}
 & \arbor{} & 78\% & 80\% \\
 & PRM & 9\% & 9\% \\
 & token conf. & 11\% & 11\% \\
\bottomrule
\end{tabular}
\caption{\textbf{Expert review confirms \arbor{} and rejects local signals.} Precision: share of a method's flagged steps the annotator confirms as the stated failure. Recall: share of the confirmed steps of the failure mode (all methods' steps pooled) that the method also flags. For underthinking we also showed the annotators the lineage \arbor{} tracked, since the text rarely settles whether the abandoned branch was still viable (Appendix~\ref{sec:app-local-annotation}).}
\label{tab:local-expert}
\end{table}

\paragraph{Local signals.}
We compare \arbor{} with the local signals the literature uses to detect the same failures: for repetition, the \emph{n-gram} rule of \citet{pipis2026loop} and \emph{embedding similarity} between adjacent sentence windows (IRD; \citealp{zhen2025reconsidering}); for continuing after the answer, \emph{post-answer redundancy} (ERD; \citealp{zhen2025reconsidering}); for underthinking, \emph{switch tokens} \citep{wang2025underthinking}; and for unsupported deduction, a \emph{process reward model} (PRM, Qwen2.5-Math-PRM-7B; \citealp{zhang2025lessons}) and \emph{token confidence} \citep{fu2025deepconf}, each flagging a trajectory's lowest-scored steps.
We set each threshold on development trajectories so that the signal flags the same share of tokens as \arbor{}, which compares precision at equal coverage (Appendix~\ref{sec:app-local-tuning}).

\paragraph{Experts confirm \arbor{} and reject the local signals.}
Four annotators confirm most of \arbor{}'s steps and reject most of the local signals', and \arbor{} also recovers more of the confirmed steps (Table~\ref{tab:local-expert}).
For repetition the two are comparable, because \arbor{} records part of what the annotators count as repetition as a restatement rather than a re-derivation.

\paragraph{Local signals flag ordinary reasoning.}
The local signals recover only a small share of \arbor{}'s steps in every failure mode (Figure~\ref{fig:local-coverage}).
Most of the steps they do flag are ordinary reasoning, first derivations or verifications and discourse such as ``Wait'' or ``Let me see'' (Figure~\ref{fig:local-composition}).
Moreover, half of the steps that switch tokens flag as underthinking actually leave a branch the oracle had already ruled out, so a penalty on switch tokens would suppress justified backtracking as often as underthinking (Table~\ref{tab:app-local-switch}).

\begin{figure}[t]
    \centering
    \includegraphics[width=\linewidth]{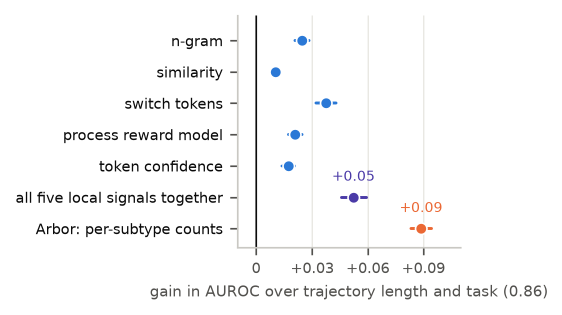}
    \caption{\textbf{\arbor{}'s subtype counts predict the outcome better than all local signals combined.} Gain in AUROC for separating incorrect from correct trajectories over a baseline of trajectory length and task (solid line, 0.86), pooled over the three tasks. Error bars: 95\% confidence intervals.}
    \label{fig:local-auroc}
\end{figure}

\paragraph{\arbor{}'s subtype counts separate incorrect from correct trajectories better than all local signals combined.}
\arbor{}'s per-subtype counts separate incorrect from correct trajectories better than any single local signal and better than the five of them together (Figure~\ref{fig:local-auroc}), and adding the five local signals to \arbor{}'s does not further increase AUROC (per-task and length-normalized results in Appendix~\ref{sec:app-local}).
This indicates that the local signals only carry a subset of what \arbor{} encodes.
\section{Cost and Mechanism of Each Failure Mode}
\label{sec:results-subtypes}
\label{sec:results-subtypes-setup}

\begin{table}[t]
\centering
\footnotesize
\setlength{\tabcolsep}{2pt}
\begin{adjustbox}{max width=\linewidth}
\begin{tabular}{@{}l rr rr@{}}
\toprule
& \multicolumn{2}{c}{Token cost (\%)} & \multicolumn{2}{c}{OR of ending} \\
\cmidrule(lr){2-3} \cmidrule(lr){4-5}
Subtype & self & ext. & incorr. & trunc. \\
\midrule
\multicolumn{5}{@{}l}{\textit{Repetition}} \\
\quad R1 Re-Derivation & 1.0 & 0.3 & 0.98 & 0.82 \\
\quad R2 Derive-after-Prune & 1.0 & 0.4 & 0.66 & \textbf{1.42} \\
\quad R3 Redundant Re-Verification & \textbf{1.2} & 0.3 & 0.37 & 0.45 \\
\quad R4 Subgoal Churn & 0.2 & 0.3 & \textbf{1.28} & \textbf{1.15} \\
\midrule
\multicolumn{5}{@{}l}{\textit{Underthinking}} \\
\quad R5 Branch Abandonment & 0.3 & 20.8 & \textbf{1.07} & 0.64 \\
\quad R6 Subgoal Abandonment & 0.06 & \textbf{46.7} & \textbf{1.24} & \textbf{1.18} \\
\quad R7 Post-Answer Attempt & 0.2 & 0.7 & 0.08 & 0.77 \\
\midrule
\multicolumn{5}{@{}l}{\textit{Unsupported deduction}} \\
\quad R8 Premature Finalization & 0.04 & 4.9 & \textbf{10.12} & 0.78 \\
\quad R9 Infeasible Derivation & \textbf{1.2} & 2.6 & \textbf{2.90} & \textbf{1.83} \\
\quad R10 False Rejection & 0.2 & 37.5 & 0.87 & 1.00 \\
\quad R11 False Verdict & 0.4 & 0.4 & \textbf{1.70} & 0.91 \\
\bottomrule
\end{tabular}
\end{adjustbox}
\caption{\textbf{The cost of each subtype} (Qwen3-8B, pooled over the three tasks; measurements as defined in \S\ref{sec:results-subtypes-setup}). OR: odds ratio. Per-task values, intervals and significance in Table~\ref{tab:subtypes}.}
\label{tab:subtypes-main}
\end{table}

\begin{figure*}[t]
    \centering
    \includegraphics[width=\textwidth]{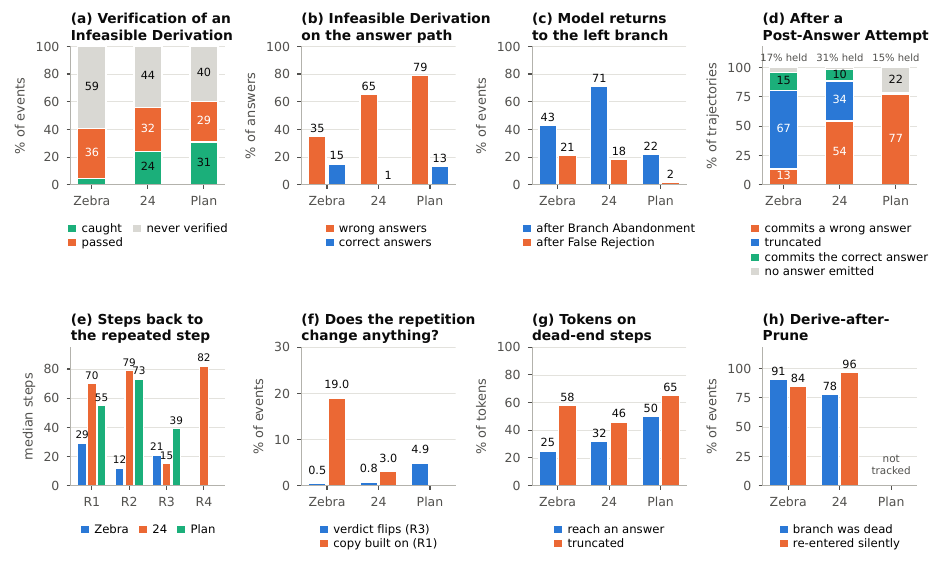}
    \caption{\textbf{The evidence behind \S\ref{sec:results-subtypes}}, per task (Qwen3-8B; exact values in Tables~\ref{tab:app-answer} to~\ref{tab:app-budget}). In (d), the label above each bar is the share of failing trajectories that held a correct solution; in (h), PlanBench has no branch lifecycle.}
    \label{fig:subtype-evidence}
\end{figure*}

For each failure mode, we ask two questions.
\textbf{Q1}: How does it harm reasoning, in token cost, in answer correctness, or both?
\textbf{Q2}: What mechanism lets it arise or persist?

We use the following measurements in our analysis, detailed below:
(1) \emph{Self cost}: the share of all tokens on the steps that carry the subtype.
(2) \emph{Extended cost}: the share of a trajectory's tokens spent in the region of the tree that an event of the subtype opens, defined per subtype (for an abandonment, the steps generated outside the abandoned branch until the model returns to it; for an unsupported deduction, the subtree rooted at the offending node), as the median over events.
(3) \emph{Incorrect and truncation odds ratios}: how much likelier a trajectory containing the subtype is to end incorrect rather than correct, or truncated rather than completed, than a trajectory of the same length on the same task.
The main text reports only the values that characterize each subtype; Appendix~\ref{sec:app-subtypes} gives every statistic for every subtype and task.

\subsection{Unsupported Deduction}
Unsupported deduction is the failure mode that costs correctness: the model's own verification seldom catches a step the state does not support, and the committed answer builds on it.
Three of its four subtypes have the largest incorrect odds ratios of any subtype, at small extended cost (Table~\ref{tab:subtypes-main}).
Verification catches an \textbf{Infeasible Derivation} in under a third of cases on every task (Figure~\ref{fig:subtype-evidence}a).
\textbf{Infeasible Derivation} lies on the dependency path of most wrong answers and few correct ones on every task (Figure~\ref{fig:subtype-evidence}b).

The exception is \textbf{False Rejection}, which costs tokens rather than correctness: verification wrongly rejects a viable branch, and the model rarely returns to a branch it has rejected.
Its incorrect odds ratio is below 1 but its extended cost is 37.5\% of the trajectory (Table~\ref{tab:subtypes-main}).
The model returns to a rejected branch far less often than to an abandoned one on every task (Figure~\ref{fig:subtype-evidence}c).

\subsection{Underthinking}
Underthinking costs tokens rather than correctness: the model typically still reaches the answer, whether by returning to the abandoned branch or by an alternative path, and the cost of underthinking is the search expended in between.
Branch and \textbf{Subgoal Abandonment} have the two largest extended costs of any subtype with incorrect odds ratios near 1 (Table~\ref{tab:subtypes-main}).
The model returns to the abandoned branch in between a fifth and seven tenths of cases, depending on the task (Figure~\ref{fig:subtype-evidence}c).

The exception is \textbf{Post-Answer Attempt}, which costs correctness: the model leaves a complete correct solution rather than an unfinished branch, and rarely returns to it.
Between a sixth and a third of failing trajectories held a correct solution, and after the attempt most of them commit a wrong answer or run out of budget (Figure~\ref{fig:subtype-evidence}d).
Its incorrect odds ratio of 0.08 reflects that most such attempts occur in trajectories that end correct (Table~\ref{tab:subtypes-main}).

\subsection{Repetition}
Repetition does less harm to the answer than the other two modes and costs few tokens beyond its own steps: the model re-establishes a step it had proposed long before in the trajectory, and the re-establishment adds nothing to what that step had already given.
Its subtypes have the largest self costs but the smallest extended costs of any subtype, and incorrect odds ratios below 1 except for \textbf{Subgoal Churn} (Table~\ref{tab:subtypes-main}).
The re-established step lies tens of steps back for every subtype and task (Figure~\ref{fig:subtype-evidence}e).
A re-verification flips the verdict in under 5\% of cases, and later work rarely builds on a \textbf{Re-Derivation}'s copy (Figure~\ref{fig:subtype-evidence}f).

The budget that repetition does not consume goes to dead ends: truncated trajectories spend more of their tokens on steps that can no longer reach a correct solution than trajectories that reach an answer do, on every task (Figure~\ref{fig:subtype-evidence}g).
The exception within repetition is \textbf{Derive-after-Prune}, the one subtype associated with truncation (Table~\ref{tab:subtypes-main}): the model re-enters a branch that was in fact dead, without re-opening it, in the large majority of cases (Figure~\ref{fig:subtype-evidence}h).

\section{Precursors of Failure in the Tree State}
\label{sec:results-precursors}
\label{sec:results-precursors-setup}

\paragraph{Setup.}
Section~\ref{sec:results-subtypes} established what each failure costs and how it arises; a mitigation, however, must act before the failure.
We therefore ask whether the tree state before a step already tells a failure apart from the steps at which it could have occurred but did not, which we call \emph{at-risk steps}.
For each subtype we take the steps at which it occurred and, as controls, at-risk steps, on which its precondition held, matched on task, operation and position.
We describe each step by the tree state and the operations over the preceding steps and train a classifier to separate the two sets on held-out problems, against two baselines, position alone and the previous operation alone (Appendix~\ref{sec:app-precursors}).

\begin{table}[t]
\centering
\footnotesize
\setlength{\tabcolsep}{3pt}
\begin{adjustbox}{max width=\linewidth}
\begin{tabular}{@{}l l rr@{}}
\toprule
& & \multicolumn{2}{c}{AUROC} \\
\cmidrule(lr){3-4}
Kind of state & Subtype & State & Labels \\
\midrule
\multirow{4}{*}{Branch content}
 & R1 Re-Derivation & 0.73 & 0.70 \\
 & R2 Derive-after-Prune & 0.59 & 0.51 \\
 & R4 Subgoal Churn & 0.77 & 0.64 \\
 & R9 Infeasible Derivation & 0.78 & 0.71 \\
\midrule
Current node verified? & R5 Branch Abandonment & 0.91 & 0.81 \\
\midrule
\multirow{2}{*}{Recent operations}
 & R3 Redundant Re-Verification & 0.83 & 0.81 \\
 & R7 Post-Answer Attempt & 0.87 & 0.87 \\
\bottomrule
\end{tabular}
\end{adjustbox}
\caption{\textbf{What the precursor of each subtype reads, and whether the tree is needed to read it} (24~Game). State: held-out AUROC of a classifier on the full tree state over the preceding steps, separating the steps at which the subtype occurred from matched at-risk steps, at which it could have occurred but did not. Labels: the same classifier given only the operation label of each preceding step, such as derive or check, and the position in the trajectory. The other tasks and the position and previous-operation baselines are in Table~\ref{tab:precursors}.}
\label{tab:precursor-families}
\end{table}

\paragraph{The tree state is crucial for separating a failure from its at-risk steps.}
For each subtype, the stable precursor is the set of state features the classifier selects in every refit, stated as a rule in Table~\ref{tab:app-precursor-rules}.
These rules fall into three kinds of state (Table~\ref{tab:precursor-families}).
The first is \emph{branch content}, the constraints fixed, assumptions opened and nodes created on the branch; the second is \emph{whether the current node has been verified}; the third is \emph{recent operations}, the last few operation labels.
The tree adds information beyond the operation labels and position: a classifier refit on labels and position alone separates the subtypes whose precursor is recent operations as well as the full state does, but falls clearly short on those whose precursor is branch content or the verification of the current node (Table~\ref{tab:precursor-families}), and these latter two kinds precede the subtypes that build wrong answers or leave viable work behind.
Table~\ref{tab:precursor-families} reports the 24~Game, where the state separates every modeled subtype well above both baselines; on ZebraLogic and PlanBench the separation is weaker and strong only for \textbf{Derive-after-Prune} and \textbf{Infeasible Derivation} respectively (Table~\ref{tab:precursors}).

\paragraph{Two routes carry \arbor{}'s offline signals into model development.}
(1) At inference time, we train a lightweight predictor on \arbor{}'s labels, a probe on the model's hidden states or a monitor over the trajectory so far \citep{kortukov2026futureprobes,baker2025monitoring}, which fires when one of the three kinds of state precedes a failure.
Its signal gives existing interventions, a verification prompt, a thought-switch penalty or early exit \citep{wu2025thinkingintervention,wang2025underthinking,muennighoff2025s1}, a more accurate precondition than the unconditional one they use now; Table~\ref{tab:app-routes} lists the condition each intervention gates on.
(2) Offline, \arbor{}'s labels mark the repetition steps and dead-end subtrees that trajectory compression should remove, the repeated operation that a reinforcement-learning penalty should charge, and the failures by which sample selection should rank trajectories, in place of the token-count and token-importance criteria these methods use today \citep{xia2025tokenskip,luo2025o1pruner,hassid2025dontoverthink}.
In both routes the intervention already exists; \arbor{} supplies the condition for when to act or the label for what to keep.

\section{Conclusion}
\label{sec:conclusion}
We presented \arbor{}, a reasoning-failure attribution framework that maintains the solution state of a reasoning trajectory step by step and attributes each failure to a relation between a step and that state rather than to the step's text.
On three reasoning tasks, we showed that \arbor{} flags reasoning failures more accurately than local signals and finds the failures they miss.
We used \arbor{} to separate the subtypes of each failure mode, measure their cost in tokens and in correctness, and explain the mechanism by which each arises.
We used the tree state \arbor{} tracks to separate a failure from the at-risk steps at which it did not occur, and found that for the subtypes that cost correctness or viable work only the tree records that state.
Finally, we discussed how \arbor{}'s labels can gate inference-time interventions and replace length-based criteria in offline trajectory compression, reward shaping and sample selection.

\clearpage
\section*{Limitations}
\label{sec:limitations}
Three limitations bound the scope of our findings.
First, we developed and evaluated \arbor{} on the Qwen3 model family: we tuned the parser's schemas and retrieved examples to this family's reasoning style, and established its accuracy only on reference labels drawn from it.
Applying the framework to another model family requires re-tuning and re-validating the parser before one can trust the engine's state, and therefore every downstream count.
Second, the engine presupposes a task whose answers are verifiable and whose search space is finite, so that it can represent the state of a branch and decide its viability and the soundness of a step.
This confines the present study to tasks of that kind; open-ended reasoning, mathematical proof and agentic workflows, in which the state is not readily parseable and the notion of a viable branch is not well defined, fall outside its scope, and we leave defining engine states for them to future work.
Third, the intervention points and the mitigations built on them are untested proposals derived from the pre-state analysis.
Whether gating existing mitigations on these states, or training on trajectories curated with \arbor{}'s labels, reduces the failures in practice remains open, and we encourage future work to test it.

\section*{Ethical Considerations}
\arbor{} labels failures in model-generated reasoning on synthetic puzzle and planning tasks; it processes no personal data and makes no decisions about people.
The four expert annotators were NLP researchers who volunteered, judged only model-generated steps, and saw no personal information.

The labels are definitions over a tracked state, not ground truth about why a model reasons as it does, so a mitigation trained or gated on them inherits any error of the parser or engine; we report the parser's accuracy and the engine's validation so that users can weigh that error.
Reducing repetition or switching on the basis of these labels could suppress useful deliberation if applied beyond the tasks on which we validated them, and a monitor built on the labels should not be read as evidence about a model's reasoning outside those tasks.

The environmental cost of this work is the GPU inference for generating and parsing trajectories (\S\ref{sec:parser}, Appendix~\ref{sec:setup}); the engine and detectors run on a CPU. Parsing adds one LLM call per step, a cost a deployed monitor would also incur.

\section*{Release and License}
We release the parser schemas and prompts, the engine, the detector definitions and the analysis scripts at \url{https://github.com/INK-USC/ARBOR} under the MIT license, and the full parsed trajectories, their trees and the analysis traces behind every result at \href{https://doi.org/10.5281/zenodo.23405752}{doi:10.5281/zenodo.23405752} under CC BY 4.0.
The project page (\url{https://inklab.usc.edu/ARBOR}) hosts interactive examples of trajectories and their trees.

\section*{Use of AI Assistants}
Large language models are part of the method: a Qwen3 model parses each step (\S\ref{sec:parser}), and LLM judges re-labeled the reference trajectories and validated engine decisions under written rules with human adjudication (Appendices~\ref{sec:appendix_parser} and~\ref{sec:appendix_validation}).
We also used an AI assistant to edit the text of this paper and to write analysis scripts; the authors reviewed every edit and every result.

\bibliography{custom}

\appendix

\section{Related Work}
\label{sec:related}

\paragraph{Local proxies for the three failure modes.}
Prior work detects repetition as verbatim looping, by the rule that a response loops if some $n$-gram recurs at least $k$ times \citep{pipis2026loop}, or as embedding similarity between nearby sentence windows \citep{zhen2025reconsidering}. It counts overthinking as the solution rounds after the first correct answer \citep{chen2025overthinking}, and underthinking as the token share spent after the first correct step is abandoned and the frequency of thought-switch tokens \citep{wang2025underthinking}. Erroneous steps are the domain of process supervision, with step labels from human annotation \citep{lightman2024verify}, Monte-Carlo rollouts \citep{wang2024mathshepherd,zhang2025lessons} or hidden-state scores \citep{sun2026rhd}, and benchmarks that ask a verifier for the first wrong step \citep{zheng2025processbench}. All of these signals are defined over the text of a step or of the whole trace, and none records how a step depends on earlier steps.

\paragraph{Structured views of reasoning traces.}
Trees and graphs over steps first served to generate reasoning under explicit search \citep{yao2023tot}. For analysis, prior work segments a trace at cue words such as \emph{wait} or \emph{alternatively} and labels each block by its function, such as execution, reflection or transition \citep{chen2025seal,wang2026segmentattribution}, or an LLM maps each trace step to a step of an extracted sketch \citep{jiang2025lcot2tree}, to a successful or failed attempt \citep{feng2026fsf} or to a node of a dependency graph \citep{yuan2026graphpruning}. A separate strand clusters hidden states into graph nodes that carry no symbolic content \citep{minegishi2025topology}. In each case the structure comes from the wording of a block or from a judge, and it records what a block looks like rather than what it does to the solution state.

\paragraph{Proxies as the basis for interventions.}
The proxies above directly drive mitigations: training-side compression and selection by solution-round counts, budget forcing or perplexity criteria \citep{chen2025overthinking,muennighoff2025s1,cui2025spirit,wang2026segmentattribution}, RL penalties on redundancy or hidden-state scores \citep{zhen2025reconsidering,sun2026rhd}, and decoding-side suppression of switch tokens, repeated chunks or reflection states, early exit, and pruning of review nodes \citep{wang2025underthinking,xie2025wordsalad,wei2026evolution,chen2025seal,yuan2026graphpruning}. Each intervention carries an implicit label, that repeated or post-answer text is useless or that a switch is harmful, which no method checks at the individual step, and the accuracy lost on removing pre-answer redundancy \citep{zhen2025reconsidering} is one sign that the label is too coarse.

\FloatBarrier

\section{Experimental Details}
\label{sec:setup}

\subsection{Tasks and datasets}

We use three tasks with distinct reasoning structures.
\textbf{ZebraLogic} \citep{lin2025zebralogic}: constraint satisfaction over a
row of houses; we use the 1{,}000 \texttt{grid\_mode} puzzles (25 grid sizes
from $2{\times}2$ to $6{\times}6$, 40 puzzles each), each with a unique
solution.\footnote{\texttt{WildEval/ZebraLogic}.}
\textbf{24 Game}: search over arithmetic compositions; we use the 1{,}362
solvable puzzles of the train split (the dataset has no test
split).\footnote{\texttt{nlile/24-game}.}
\textbf{PlanBench} \citep{valmeekam2023planbench}: sequential action planning
under preconditions, one-shot plan generation.\footnote{\texttt{tasksource/planbench},
\texttt{task\_1\_plan\_generation}.} We generated for the first
1{,}664 of the benchmark's 2{,}270 questions, and every analysis uses that
set: Blocksworld 500, Depots 500, Obfuscated-Deceptive Logistics 285,
Mystery Blocksworld 179, Mystery Blocksworld-3 100 and Blocksworld-3 100;
the set does not cover plain Logistics. All three benchmarks are public and we use them as released, for research.

\subsection{Models and generation}

We study Qwen3 at three sizes --- 1.7B, 8B and 32B --- in thinking mode;
the 8B model is the main corpus and the other two sizes serve the
size comparison. We send each question as a single user message through
the model's chat template with thinking enabled, and generate with
\texttt{vLLM} at a fixed seed of 42. ZebraLogic and 24 Game use
temperature 0.7, top-$p$ 1.0, $n=3$ samples per question and a maximum of
8{,}192 generated tokens. PlanBench uses the model's recommended thinking
settings --- temperature 0.6, top-$p$ 0.95, top-$k$ 20 --- and a maximum of
16{,}000 tokens; we drew eight samples per question at 1.7B and 8B and
three at 32B, and use only the first three anywhere. Every model
receives the same task-specific prompt templates
(Appendix~\ref{sec:appendix_prompts}). This gives 3{,}000 ZebraLogic,
4{,}086 24~Game and 4{,}992 PlanBench trajectories per model size. We
exclude from every number in the main text the 40 questions per task on
which we developed the parser and engine
(Appendix~\ref{sec:appendix_parser_reference}),
leaving 2{,}880, 3{,}966 and 4{,}872 trajectories at 8B. The size
comparison uses a panel of 90 questions per task, balanced
over the 8B outcome (solved in 3, 1--2, or 0 of its samples) and over
length, and run at all three sizes. We ran all generation and parsing on NVIDIA RTX A6000, RTX 8000 and A100 GPUs. Qwen3 is released under the Apache-2.0 license.

\subsection{Evaluation}

A deterministic, task-specific rule scores each trajectory on the
answer that follows the thinking. A trajectory that reaches the token
limit without an answer is \emph{truncated}; one whose answer cannot be
read in the required format is \emph{invalid}; the rest are correct or
incorrect.
\textbf{ZebraLogic} expects a JSON grid; it is correct when every house's
attributes equal the unique solution. \textbf{24 Game} expects
\texttt{ANSWER: <expression>}; the expression must use exactly the four
given numbers (otherwise invalid) and is correct when it evaluates to 24;
\texttt{NO\_SOLUTION} is incorrect since every puzzle is solvable.
\textbf{PlanBench} expects one action per line ending in
\texttt{[PLAN END]}; we score \emph{plan validity} --- we replay the plan
from the initial state, and it is correct when every action is applicable and
the goal holds at the end --- rather than exact match to the benchmark's
single gold plan, since many valid plans exist (at 8B, 25.6\% of
trajectories are valid against 15.2\% exact matches; at 32B, 57.1\%
against 27.5\%). Table~\ref{tab:main_results}
reports the outcome shares on the full corpora.

\begin{table}[t]
\centering
\small
\setlength{\tabcolsep}{4pt}
\caption{Outcome shares (\%) per task and model size over the first three samples of every question (ZebraLogic 3{,}000, 24~Game 4{,}086, PlanBench 4{,}992 trajectories per size). PlanBench ``correct'' is plan validity; ``invalid'' is no plan emitted.}
\label{tab:main_results}
\begin{tabular}{@{}ll ccc@{}}
\toprule
\textbf{Task} & \textbf{Model} & Correct & Truncated & Invalid \\
\midrule
\multirow{3}{*}{ZebraLogic} & Qwen3-1.7B & 49.0 & 39.6 & 0.0 \\
 & Qwen3-8B & 70.2 & 27.0 & 0.0 \\
 & Qwen3-32B & 75.4 & 22.9 & 0.0 \\
\midrule
\multirow{3}{*}{24 Game} & Qwen3-1.7B & 80.8 & 6.4 & 12.1 \\
 & Qwen3-8B & 78.0 & 7.0 & 10.8 \\
 & Qwen3-32B & 78.8 & 5.7 & 14.8 \\
\midrule
\multirow{3}{*}{PlanBench} & Qwen3-1.7B & 2.6 & 0.4 & 3.3 \\
 & Qwen3-8B & 25.6 & 0.6 & 0.1 \\
 & Qwen3-32B & 57.1 & 2.4 & 0.0 \\
\bottomrule
\end{tabular}

\end{table}

\FloatBarrier
\section{Model size}
\label{sec:app-scale}
\begin{figure*}[t]
\centering
\includegraphics[width=\linewidth]{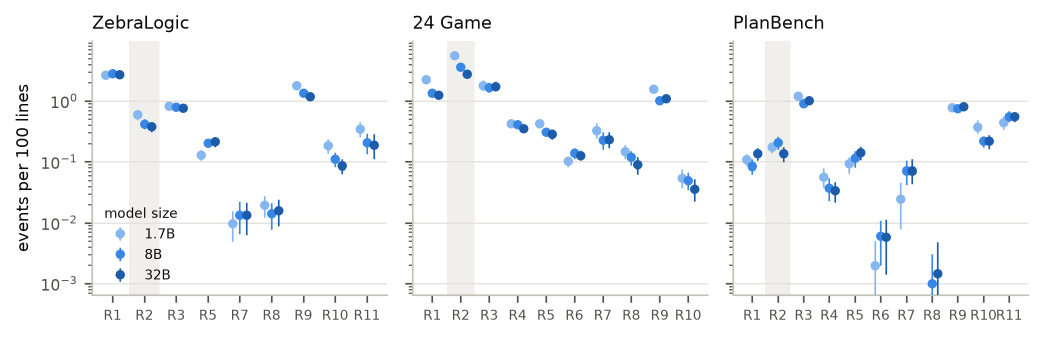}
\caption{\textbf{Subtype rates by model size.} For each task and subtype, the number of events per 100 steps (log scale) at 1.7B, 8B and 32B, left to right within each group, with 95\% intervals over problems, on the same 86, 88 and 80 questions per task at every size. A group whose three dots descend is a subtype that becomes rarer with size; most groups are flat. Derive-after-Prune (R2, shaded) is the only subtype that falls at both size increases. Subtypes a task does not define are absent from its panel.}
\label{fig:subtype-scale}
\end{figure*}

We compare three sizes of the Qwen3 family, 1.7B, 8B and 32B, on the same questions, and ask whether any subtype appears or disappears with scale and how each subtype's rate changes.
Every subtype persists across the three sizes.
Scale lowers the rates of three of the four repetition subtypes and of the two unsupported-deduction subtypes that act on a single step.
The rates of underthinking and of the two subtypes that commit or judge a whole lineage, which include the subtypes associated with wrong answers, do not fall with scale.
Figure~\ref{fig:subtype-scale} and Table~\ref{tab:app-elasticity} give the rates and elasticities on the matched panel (the same 86 / 88 / 80 problems per task at 1.7B, 8B and 32B, one sampling run per size, 258 / 264 / 240 traces per task and size).
From 1.7B to 8B, the per-step rates of False Rejection, Infeasible Derivation, Derive-after-Prune, Re-Derivation and Redundant Re-Verification fall; from 8B to 32B only Derive-after-Prune keeps falling.
Branch Abandonment, Post-Answer Attempt, Premature Finalization and False Verdict do not change from 1.7B to 8B or from 8B to 32B, and Subgoal Abandonment rises from 1.7B to 8B and then holds.

\begin{table}[t]
\centering
\footnotesize
\setlength{\tabcolsep}{3pt}
\begin{adjustbox}{max width=\linewidth}
\begin{tabular}{l ll}
\toprule
Subtype & 1.7B $\to$ 8B & 8B $\to$ 32B \\
\midrule
R1 Re-Derivation & \textbf{-0.14 {\scriptsize [-0.23, -0.05]}} & 0.04 {\scriptsize [-0.03, 0.10]} \\
R2 Derive-after-Prune & \textbf{-0.45 {\scriptsize [-0.57, -0.32]}} & \textbf{-0.18 {\scriptsize [-0.30, -0.07]}} \\
R3 Redundant Re-Verification & \textbf{-0.13 {\scriptsize [-0.23, -0.02]}} & 0.04 {\scriptsize [-0.05, 0.14]} \\
R4 Subgoal Churn & -0.11 {\scriptsize [-0.29, 0.07]} & -0.05 {\scriptsize [-0.26, 0.17]} \\
R5 Branch Abandonment & -0.01 {\scriptsize [-0.15, 0.13]} & 0.06 {\scriptsize [-0.08, 0.20]} \\
R6 Subgoal Abandonment & \textbf{0.29 {\scriptsize [0.01, 0.59]}} & 0.02 {\scriptsize [-0.27, 0.29]} \\
R7 Post-Answer Attempt & -0.16 {\scriptsize [-0.42, 0.10]} & 0.03 {\scriptsize [-0.20, 0.28]} \\
R8 Premature Finalization & -0.26 {\scriptsize [-0.56, 0.03]} & -0.14 {\scriptsize [-0.45, 0.15]} \\
R9 Infeasible Derivation & \textbf{-0.29 {\scriptsize [-0.38, -0.20]}} & -0.00 {\scriptsize [-0.09, 0.09]} \\
R10 False Rejection & \textbf{-0.47 {\scriptsize [-0.69, -0.25]}} & -0.18 {\scriptsize [-0.45, 0.05]} \\
R11 False Verdict & -0.07 {\scriptsize [-0.35, 0.22]} & -0.09 {\scriptsize [-0.32, 0.12]} \\
\bottomrule
\end{tabular}
\end{adjustbox}
\caption{Elasticity of each subtype's rate with model size on the pooled panel: the log ratio of rates over the log ratio of parameter counts, with 95\% cluster-bootstrap intervals over problems. Bold: interval excludes 0. Only R2 is negative on both steps; no subtype is positive on both.}
\label{tab:app-elasticity}
\end{table}

\FloatBarrier
\section{Prompts}
\label{sec:appendix_prompts}

We use the three templates below unchanged for every model size; \texttt{\{question\_text\}} is the puzzle. For PlanBench the puzzle is the benchmark's own one-shot query (domain description, a worked example, and the target instance), shown here for a Blocksworld instance; only the final instruction line is ours.

\begin{table*}[htbp]
\centering
\begin{tcolorbox}[
    colback=gray!3!white,
    colframe=gray!50!black,
    boxrule=0.5pt,
    width=15cm,
    arc=2mm,
    title=\text{Prompt Template for 24 Game},
    fonttitle=\bfseries,
    coltitle=white,
]
\texttt{You are a 24-game solver.}

\vspace{0.5em}
\texttt{GOAL}\\
\texttt{Given 4 numbers (TILES), build an expression that equals exactly 24 using "+", "-", "*", "/" and parentheses.}\\
\texttt{Use each tile exactly once. Intermediate fractions are allowed. No concatenation (e.g., 1 and 2 cannot form 12). Avoid division by zero. Use standard operator precedence unless parentheses change it.}

\vspace{0.5em}
\texttt{RULES}\\
\texttt{- Use every tile exactly once.}\\
\texttt{- Only "+", "-", "*", "/" and parentheses may appear.}\\
\texttt{- No division by zero.}\\
\texttt{- The value must equal exactly 24.}

\vspace{0.5em}
\texttt{REQUIRED VERIFICATION}\\
\texttt{Before the final answer, verify:}\\
\texttt{- All tiles are used exactly once.}\\
\texttt{- The evaluated value equals 24.}

\vspace{0.5em}
\texttt{End response with one of the following exactly:}\\
\texttt{ANSWER: <fully parenthesized valid expression evaluating to 24>}\\
\texttt{or}\\
\texttt{ANSWER: NO\_SOLUTION}

\vspace{0.5em}
\texttt{Example:}

\vspace{0.5em}
\texttt{\hspace{0em}Given:}\\
\texttt{\hspace{2em}Tiles: [3, 3, 8, 8]}\\
\texttt{\hspace{2em}Target: 24}

\vspace{0.5em}
\texttt{\hspace{0em}Solution:}\\
\texttt{\hspace{2em}8 / (1/3) = 24}\\
\texttt{\hspace{2em}(1/3) = (3 - (8/3))}\\
\texttt{\hspace{2em}Full expression = (8/(3-(8/3))) = 24}

\vspace{0.5em}
\texttt{\hspace{0em}Verification:}\\
\texttt{\hspace{2em}Final expression: (8/(3-(8/3)))}\\
\texttt{\hspace{2em}required count: 3:2, 8:2}\\
\texttt{\hspace{2em}final expression count: 3:2, 8:2}\\
\texttt{\hspace{2em}Every tile used exactly once: yes}\\
\texttt{\hspace{2em}Evaluate: (8/(3-(8/3))):}\\
\texttt{\hspace{2em}3 - 8/3 = 1/3, 8 / (1/3) = 24}\\
\texttt{\hspace{2em}Final value equals 24: yes}

\vspace{0.5em}
\texttt{\hspace{0em}ANSWER: (8/(3-(8/3)))}

\vspace{0.5em}
\texttt{INPUT}\\
\texttt{TILES: \{question\_text\}}
\end{tcolorbox}
\captionof{table}{Prompt Template for 24 Game}
\label{tab:prompt_24game}
\end{table*}

\begin{table*}[htbp]
\centering
\begin{tcolorbox}[
    colback=gray!3!white,
    colframe=gray!50!black,
    boxrule=0.5pt,
    width=15cm,
    arc=2mm,
    title=\text{Prompt Template for Zebra Logic},
    fonttitle=\bfseries,
    coltitle=white,
]
\texttt{\# Example Puzzle}

\vspace{0.5em}
\texttt{There are 3 houses, numbered 1 to 3 from left to right, as seen from across the street.}\\
\texttt{Each house is occupied by a different person.}\\
\texttt{Each house has a unique attribute for each of the following characteristics:}\\
\texttt{- Each person has a unique name: peter, eric, arnold}\\
\texttt{- Each person has a unique favorite drink: tea, water, milk}

\vspace{0.5em}
\texttt{\#\# Clues for the Example Puzzle}\\
\texttt{1. Peter is in the second house.}\\
\texttt{2. Arnold is directly left of the one who only drinks water.}\\
\texttt{3. The one who only drinks water is directly left of the person who likes milk.}

\vspace{0.5em}
\texttt{\#\# Reasoning}\\
\texttt{Given Clue 1, we know Peter is in House 2. According to Clue 2, Arnold is directly left of the one who only drinks water. The person in House 3 cannot be on the left of anyone, so Arnold must be in House 1. Thus, Peter drinks water, and Eric lives in House 3. Then, according to Clue 3, Eric drinks milk. Therefore, Arnold drinks tea.}

\vspace{0.5em}
\texttt{\#\# Answer to the Example Puzzle}\\
\texttt{\{}\\
\texttt{\hspace{1em}"House 1": \{"Name": "arnold", "Drink": "tea"\},}\\
\texttt{\hspace{1em}"House 2": \{"Name": "peter", "Drink": "water"\},}\\
\texttt{\hspace{1em}"House 3": \{"Name": "eric", "Drink": "milk"\}}\\
\texttt{\}}

\vspace{0.5em}
\texttt{Return the final response in the following format:}\\
\texttt{\{answer\}}

\vspace{0.5em}
\texttt{\# Please solve the following puzzle and present your solution in the aforementioned json format:}

\vspace{0.5em}
\texttt{\{question\_text\}}
\end{tcolorbox}
\captionof{table}{Prompt Template for Zebra Logic}
\label{tab:prompt_zebra}
\end{table*}

\begin{table*}[htbp]
\centering
\begin{tcolorbox}[
    colback=gray!3!white,
    colframe=gray!50!black,
    boxrule=0.5pt,
    width=15cm,
    arc=2mm,
    title=\text{Prompt Template for PlanBench},
    fonttitle=\bfseries,
    coltitle=white,
]
\texttt{I am playing with a set of blocks where I need to arrange the blocks into stacks. Here are the actions I can do}

\vspace{0.5em}
\texttt{Pick up a block}\\
\texttt{Unstack a block from on top of another block}\\
\texttt{Put down a block}\\
\texttt{Stack a block on top of another block}

\vspace{0.5em}
\texttt{I have the following restrictions on my actions:}\\
\texttt{I can only pick up or unstack one block at a time.}\\
\texttt{I can only pick up or unstack a block if my hand is empty.}\\
\texttt{I can only pick up a block if the block is on the table and the block is clear. A block is clear if the block has no other blocks on top of it and if the block is not picked up.}\\
\texttt{I can only unstack a block from on top of another block if the block I am unstacking was really on top of the other block.}\\
\texttt{I can only unstack a block from on top of another block if the block I am unstacking is clear.}\\
\texttt{Once I pick up or unstack a block, I am holding the block.}\\
\texttt{I can only put down a block that I am holding.}\\
\texttt{I can only stack a block on top of another block if I am holding the block being stacked.}\\
\texttt{I can only stack a block on top of another block if the block onto which I am stacking the block is clear.}\\
\texttt{Once I put down or stack a block, my hand becomes empty.}\\
\texttt{Once you stack a block on top of a second block, the second block is no longer clear.}

\vspace{0.5em}
\texttt{[STATEMENT]}\\
\texttt{As initial conditions I have that, the red block is clear, the blue block is clear, the yellow block is clear, the hand is empty, the blue block is on top of the orange block, the red block is on the table, the orange block is on the table and the yellow block is on the table.}\\
\texttt{My goal is to have that the orange block is on top of the blue block.}

\vspace{0.5em}
\texttt{My plan is as follows:}

\vspace{0.5em}
\texttt{[PLAN]}\\
\texttt{unstack the blue block from on top of the orange block}\\
\texttt{put down the blue block}\\
\texttt{pick up the orange block}\\
\texttt{stack the orange block on top of the blue block}\\
\texttt{[PLAN END]}

\vspace{0.5em}
\texttt{[STATEMENT]}\\
\texttt{As initial conditions I have that, the red block is clear, the yellow block is clear, the hand is empty, the red block is on top of the blue block, the yellow block is on top of the orange block, the blue block is on the table and the orange block is on the table.}\\
\texttt{My goal is to have that the orange block is on top of the red block.}

\vspace{0.5em}
\texttt{My plan is as follows:}

\vspace{0.5em}
\texttt{[PLAN]}

\vspace{0.5em}
\texttt{Provide the plan that achieves the goal, one action per line, in the same format as the example above. End your plan with [PLAN END].}
\end{tcolorbox}
\captionof{table}{Prompt Template for PlanBench (the benchmark's one-shot query for a Blocksworld instance, followed by our instruction line)}
\label{tab:prompt_planbench}
\end{table*}

\FloatBarrier
\section{Parser: adjudicated reference labels, metric, and development}
\label{sec:appendix_parser}

This appendix documents the parser of \S\ref{sec:parser}: how we built the adjudicated reference labels we measure it against, what the tolerant metric counts, how we reached the final configuration, and what we tried and did not adopt. Every number below re-scores stored parser outputs against the frozen reference and is reproducible from the released files.

\subsection{Adjudicated reference labels}
\label{sec:appendix_parser_reference}

The reference is 40 trajectories per task (120 trajectories, 46{,}984
steps), drawn from the Qwen3-8B generations by stratified selection over
outcome and length. Hand-labeling it was not feasible, and the label
scheme could not be fixed in advance, so we built it the way qualitative
coding builds a codebook, with model annotators in the loop
\citep{kim2026guidelines,he2026experts}: a stronger model labeled every step, then LLM judges re-labeled it independently from the written rules alone; where two judges agreed the label stood, and we raised every remaining disagreement to a human as a decision card quoting the steps and the candidate rules. We wrote each ruling into the rulebook, applied it corpus-wide by script rather than as an edit to a single step, and re-froze the reference with a content hash after each round. A final audit read every
step the parser gets wrong in all five of its samples against the
rulebook and found the reference right on 90\% of them. Adjudication against explicit rules, rather than inter-annotator agreement, thus decides correctness \citep{liu2026agreement}. The 40 trajectories
also served as the development set for every choice below; a held-out
review on trajectories the development never saw
\citep{pangakis2023validation} is in progress, and we report the configuration and the alternatives we tried so that the accuracy is not the best of many undisclosed configurations \citep{baumann2025hacking}.

\subsection{Labeling protocol}
\label{sec:appendix_parser_protocol}

The parser is one open-weight model, Qwen3.8-27B, with constrained JSON output. It labels steps in groups of fifteen and sees the fifteen preceding steps with the labels it gave them, a few labeled reference steps retrieved from other trajectories of the same task, and the schema elements that each step's words name. It never sees a summary of the solution state; what has been \emph{concluded} is the engine's job, so a label depends only on the sentence and its neighborhood. We take five independent samples per trajectory and decide each step's label by majority vote.

\subsection{What the tolerant metric counts}
\label{sec:appendix_parser_metric}

We compute three metrics per step over the steps of the 40 traces that carry a reference label, excluding the few steps whose reference is itself contested.

\paragraph{Strict.} The parser's label equals the reference label.

\paragraph{Tolerant.} Strict, or the pair of labels is one that the
rulings declared indistinguishable for the engine \emph{and} the
payloads agree on the facts or the action they name. The pairs are:
\begin{itemize}
\item a fact stated as a fresh inference (\textsc{Derive}/floating)
versus restated from the givens (\textsc{Restate}) versus reported as
holding (\textsc{Verify}, outcome hold), when both payloads name the
same fact --- the engine attaches the same fact to the same node either
way;
\item a plan step adopted (\textsc{Assume}) versus echoed
(\textsc{Restate}) when both name the same action; the echo test looks back over the whole current attempt, which a windowed parser cannot see;
\item a bare applicability verdict read as enabling an action
(\textsc{Derive}/action-enabled) versus as a check that holds
(\textsc{Verify}) on the same action;
\item a plan step under a visible answer header read as confirmed
(\textsc{Confirm}) versus recapped (\textsc{Restate}), and any step
label when the header lies outside the parser's window;
\item a bare closer (``Yes.'', ``So no.'', ``Met.'') read as narration
(\textsc{Meta}) versus as a payload-free check (\textsc{Verify}), and a
\textsc{Verify} whose only content was an anchor the step does not name.
\end{itemize}
Each pair encodes a ruling that the distinction either does not change
what the engine does with the step or rests on context no step-local
parser has; strict accuracy penalizes exactly these distinctions, which
the reference itself resolves inconsistently across traces. Tolerant
accuracy therefore measures whether the parser produced the label the
engine needs.

\paragraph{Payload (\textsc{Key}).} On steps whose reference payload
carries schema atoms, the set of atoms the parser emits equals the set
the reference carries: the role (kind or target), the facts with their
polarity and arguments, the actions with their arguments, and the
outcome. An anchor the step does not itself name (an action a subgoal
serves, a step a check is about) is excluded from both sides, since it
can only come from context. Cited premises are support, not key content.

\subsection{Development against the frozen reference}
\label{sec:appendix_parser_development}

We ran every candidate change as five independent samples on the frozen reference (sample standard deviation 0.001--0.002 tolerant), scored it locally, and adopted it only when it beat the previous configuration by at least 0.3~points on every sample. We edited prompts only through builders that record each change, and we never edited the reference to match a parser. Table~\ref{tab:parser_dev} gives the cumulative result.

\begin{table}[t]
\centering
\small
\setlength{\tabcolsep}{4pt}
\begin{adjustbox}{max width=\linewidth}
\begin{tabular}{@{}l ccc c c@{}}
\toprule
& \multicolumn{3}{c}{Tolerant} & Strict & \textsc{Key} \\
\cmidrule(lr){2-4}
Task & prompt only & + own labels, examples & + step facts & final & final \\
\midrule
PlanBench & 0.848 & 0.855 $\rightarrow$ 0.889 & \textbf{0.893} & 0.813 & 0.797 \\
ZebraLogic & 0.921 & 0.930$^{\dagger}$ & \textbf{0.937} & 0.901 & 0.784 \\
24 Game & 0.932 & \textbf{0.951} & 0.946$^{\dagger}$ (not adopted) & 0.933 & 0.848 \\
\bottomrule
\end{tabular}
\end{adjustbox}
\caption{Cumulative development on the frozen 40-trace reference; votes
of five samples unless marked $^{\dagger}$ (mean of one or two
samples). PlanBench shows own labels and examples as two steps
(0.855, 0.889); for ZebraLogic and 24~Game the two were introduced
together. We kept step facts where they added at least 0.3~points.}
\label{tab:parser_dev}
\end{table}

\paragraph{What helped and what did not.}
Four changes raised tolerant accuracy on the frozen PlanBench reference and we kept them: retrieved labeled examples from the same task, leave-one-trace-out (+3.3 points); own-label context on the preceding steps (+0.6); step facts inline with worked exemplars (+0.4, kept for PlanBench and ZebraLogic, dropped for the 24~Game, whose examples already evaluate every expression); and the reference audit with its rulings (+1.0).
Prompt cleaning over four rounds moved accuracy by 0.0 to +0.3 per round and we kept it where neutral or positive.
Five changes did not help and we did not adopt them: more examples per group (+0.16), examples oversampled toward \textsc{Verify} and \textsc{Confirm} (+0.0), step facts as a separate block ($-0.05$, strict $-1.3$), the labels of the following 15 steps ($-0.4$), and a second pass that lists every fact a step states (\textsc{Key} $-3.0$, since it adds cited premises the payload policy excludes).
An ensemble of four configurations with 16 samples gained +1.0 over the best single vote; we report it but did not ship it, since it is not a single configuration.

\paragraph{Retrieval details.} The example pool is the reference
itself. We index each reference step by TF-IDF over word unigrams and bigrams with entity names and numbers replaced by placeholders, so that ``So \textit{red} is clear now'' and ``So
\textit{blue} is clear now'' coincide. For each 15-step group we retrieve the nearest reference step to each current step from the same surface-form family (PlanBench domains with shared operators; ZebraLogic and 24~Game each one family) and from other traces only, then deduplicate and cap the set at 12 examples and 9{,}000 characters.
An example shows the two reference steps before it with their labels,
the step, and its reference output in the prompt's own format. The
gain by similarity of the retrieved example was $-3.8$, $-4.4$, and
$-4.0$ points of error rate for near-verbatim ($\geq 0.95$ cosine),
moderate ($0.70$--$0.95$), and loose ($<0.70$) matches respectively,
and $-0.7$ for steps that received no example; the parser does not
copy a twin's label. For unlabeled traces the pool is the whole
reference.

\paragraph{Step facts.} For each step, a string match against the
\textsc{Puzzle Schema} lists the entities, predicates, and operators
its words name (verb forms and common paraphrases included: ``on the
table'', ``hand is empty'', ``unstacking'') and the initial-state or
goal facts it re-states with the arguments in the right order; for the 24~Game we find and evaluate every arithmetic expression and mark it when it equals the target or uses exactly the four tiles. On the
reference, a PlanBench step with no anchors is \textsc{Meta} 87\% of
the time and a step with anchors 6\%; we append the facts to the step in angle brackets and demonstrate them in the examples.

\paragraph{What remains wrong.} On PlanBench, 8.2\% of steps are wrong in
all five samples of the final configuration, concentrated in four
confusions --- a grounded inference read as narration
(\textsc{Derive}$\to$\textsc{Meta}), a restated given read as narration
or as fresh inference, and a need read as adoption --- that did not
respond to more rules, more examples, or more step-local information.
ZebraLogic's plateau is the echo test (restated cell versus fresh
inference); 24~Game's is narration versus check. We did not train the parser or use a larger parser model.

\FloatBarrier
\section{Engine Design: Validation and Worked Examples}
\label{sec:appendix_engine}

\subsection{Resolution procedure}
\label{sec:appendix_resolution}

Every mutation of \S\ref{sec:engine} first decides which node the step is about, by one procedure:
\begin{enumerate}
\item the engine prefers an explicit anchor in the step (a named cell, expression or action);
\item otherwise it draws candidates from a bounded scope around the current step and the node the model is attending to;
\item it ranks candidates by how much of their content the step already matches;
\item it filters out candidates whose state is incompatible with the step;
\item it breaks remaining ties canonically, and revives a closed node when the model returns to it.
\end{enumerate}
Two cursors support the scope: the node the model is attending to and the node it last changed.

\subsection{Validating the engine and the detectors}
\label{sec:appendix_validation}

Two further conventions apply. We grade a False Verdict (R11) against what the model had written down, not against the truth, which separates a wrong verdict from a wrong premise. Rows are not disjoint; a detector emits every exclusion it makes beside the row with its reason.

A replay of the corpus makes on the order of $10^4$ tree decisions per
task, far more than a human can inspect. We validated the engine the
same way as the parser. On the 40 reference trajectories per task, LLM
judges re-derived the decision for sampled steps from the raw text
alone, without sight of the engine's state. Every disagreement with
the engine went to a human as a decision card and received a
written ruling. An append-only ledger keeps the rulings; we replay every
engine change against it before it ships, so the oracle only
grows. We validated the failure detectors on the same trajectories by
reading every fired and every excluded event of each row against its
definition; the human-annotation study of \S\ref{sec:results-local}
measures their precision on the full corpus.

\subsection{Worked examples}
\label{sec:appendix_examples}

The examples below are verbatim steps from the working corpus,
reproduced with their engine treatment; they illustrate the resolution
principle of \S\ref{sec:engine}. We identify puzzle instances by
trace id; tiles and target are as stated in the source
puzzle.

\paragraph{Extending the current line.} Trace \texttt{que352\_com0}
(tiles $\{2,2,3,3\}$, target 24) has already opened, among other
options, a plan for $24 = 3 \times 8$ with an open operand goal
``make~8.'' Step~\#9 —
\begin{quote}
\texttt{``Maybe if I can make 8 and then multiply by 3?''}
\end{quote}
resolves against that existing operand goal rather than opening a new
option. The decomposition it names is already a sibling in the tree,
so the engine edits that node in place and creates no new node.

\paragraph{Branching into new candidates.} Earlier in the same trace,
step~\#5 —
\begin{quote}
\texttt{``For example, 3*8, 4*6, 12+12, etc.''}
\end{quote}
proposes three unrelated decompositions of 24 in one step. The engine
resolves each independently and creates its own option node with its own
operand children, all as alternative children of the root — one parsed
step, three tree mutations.

\paragraph{Repetition resolved by state.} Step~\#50 of
the same trace proposes and immediately rejects
$(3 \times 2) \times (3 + 2) = 6 \times 5 = 30 \neq 24$, which the
engine creates and closes as a rejected exploration node. Three
steps later, step~\#53 —
\begin{quote}
\texttt{``How about (2 * 3) * (2 + 3) = 6 * 5 = 30.''}
\end{quote}
is a different string but the identical computation, evaluated in the
identical context (unattached exploration at the root, no goal on
either occasion). The engine resolves it to the already-closed node
from step~\#50 and records a repetition event rather than a new
node; state drives the decision. The
converse holds by the same rule: had the second occurrence
arisen under a different goal context, the identical computation would
be legitimate independent reuse, with no change in the
surface text. This is our central claim: grounding to lineage and
state separates true repetition from
superficially similar but substantively different reasoning.

\paragraph{Fresh start after a reframing.} Trace \texttt{que1060\_com2}
has an open, previously-worked quest to make $44/3$ from tiles
$\{4,9\}$. At step~\#82, after intervening steps
\texttt{``Another angle: let's think about fractions''} and
\texttt{``Suppose I take 9 and 4,''} the model writes
\begin{quote}
\texttt{``9*4=36.''}
\end{quote}
Although the tiles match the open quest, the engine rules this a fresh
start under the root rather than an attempt at the earlier quest. An
explicit reframing (a new-approach marker) lies between the quest's
last activity and this step, and a quest hosts new content by
context only while it remains the most recent framing.

\FloatBarrier
\section{Local measures: detail}
\label{sec:app-local}

\subsection{Matched-yield tuning}
\label{sec:app-local-tuning}

We tune each local measure on a development split of 42 traces per task, disjoint from the test traces and distinct from the 40 development questions excluded from the main text, and score it on the remaining test split.
The similarity threshold $\tau^{\ast}$ is the value whose flagged token share on the development traces is closest to the token share of our repetition rows (R1--R4) on the same traces: $\tau^{\ast}=0.934$ on ZebraLogic (777 adjacent development windows), $0.911$ on Game of 24 (817 windows) and $0.956$ on PlanBench (758 windows); the measure flags a window of sentences as redundant when the cosine similarity of its embedding to the preceding window exceeds the threshold.
The two step scorers take the bottom-$k$ share of steps that matches the token share of our unsupported-deduction rows: $k=0.010$ / $0.035$ (process reward model / token confidence) on ZebraLogic, $0.015$ / $0.035$ on Game of 24 and $0.02$ / $0.06$ on PlanBench, so the process reward model flags the lowest-scored 1 to 2\% of a trajectory's steps and token confidence the 3.5 to 6\% of steps with the lowest mean token log-probability; as trajectory scores for outcome prediction both use the mean over the worst 10\% of steps.
The $n$-gram rule scores a trajectory by the maximum number of times any 30-token $n$-gram recurs; run at its published setting (30, 20) it flags one Game of 24 trace, no ZebraLogic trace and two PlanBench traces, and relaxed to $(5, k)$ it selects 49\% first work and 48\% discourse with a repetition enrichment of 0.70, 0.00 and 1.61 per task.
The similarity scores reproduce the reference implementation to six decimals on three traces spanning Game of 24 and PlanBench.

\subsection{Expert annotation}
\label{sec:app-local-annotation}
\begin{figure*}[t]
    \centering
    \includegraphics[width=0.9\textwidth]{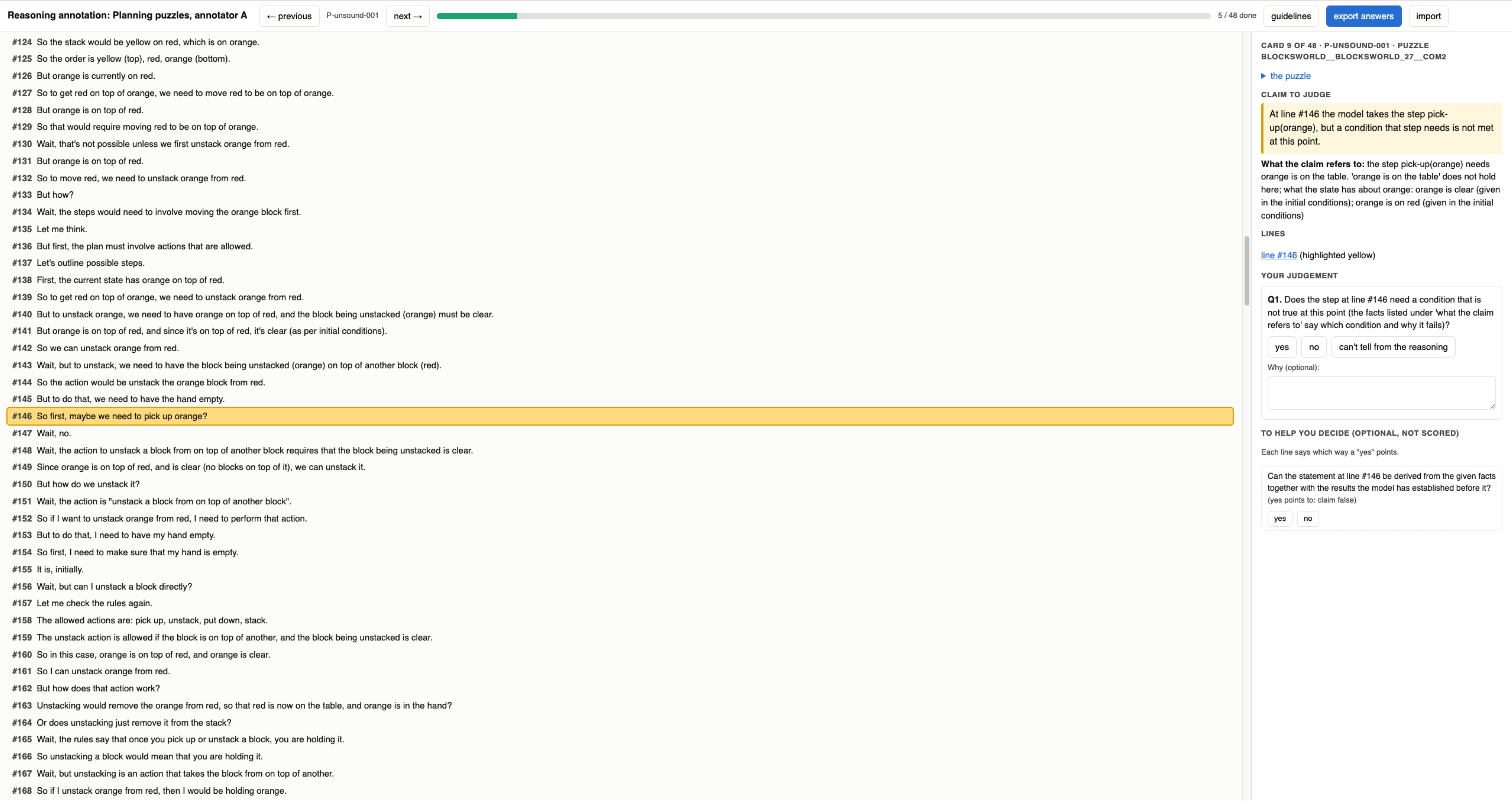}
    \caption{\textbf{The annotation interface.} One card as the annotators saw it: the flagged step in its surrounding steps, the instruction stating the failure to judge, and the response options including ``undecided''.}
    \label{fig:annotation-interface}
\end{figure*}

The four annotators are researchers in NLP with expertise in language models, who volunteered and were not paid; each read a card showing the flagged step with its surrounding steps and answered whether the step is the stated failure, with ``undecided'' allowed; Figure~\ref{fig:annotation-interface} shows the interface with its full instruction.

\begin{table}[t]
\centering
\footnotesize
\setlength{\tabcolsep}{4pt}
\begin{adjustbox}{max width=\linewidth}
\begin{tabular}{llrrrr}
\toprule
\textbf{Category} & \textbf{Method} & Cards & Conf. & Undec. & Prec. \\
\midrule
\multirow{3}{*}{Repetition} & ours (R1--R4) & 45 & 28 & 1 & 0.64 \\
 & $n$-gram & 45 & 22 & 0 & 0.49 \\
 & similarity & 45 & 28 & 0 & 0.62 \\
\midrule
\multirow{2}{*}{Abandonment} & ours (R5, R6) & 45 & 14 & 1 & 0.32 \\
 & switch marker & 45 & 12 & 1 & 0.27 \\
\midrule
\multirow{2}{*}{After the answer} & ours (R7) & 43 & 33 & 0 & 0.77 \\
 & post-answer red. & 45 & 5 & 0 & 0.11 \\
\midrule
Infeasible Derivation & ours (R9) & 34 & 27 & 0 & 0.79 \\
False Verdict & ours (R10, R11) & 11 & 8 & 0 & 0.73 \\
\midrule
\multirow{2}{*}{Wrong step, generic} & PRM & 45 & 4 & 2 & 0.09 \\
 & token confidence & 45 & 5 & 1 & 0.11 \\
\midrule
All categories & ours & 178 & 110 & 2 & 0.63 \\
\bottomrule
\end{tabular}
\end{adjustbox}
\caption{\textbf{Expert annotation, precision by category and method}, pooled over the four annotators (422 cards, each read once; a card that carries both methods' steps counts once per method). One card asks one question with the same wording for our step and for the local measure's step in the same category; a card is confirmed when the annotator answers that the step is the named failure (for abandonment, that the model leaves and the attempt was still workable). \emph{Undec.}: the annotator could not decide from the text; excluded from the precision.}
\label{tab:app-local-precision}
\end{table}

\begin{table}[t]
\centering
\footnotesize
\setlength{\tabcolsep}{4pt}
\begin{adjustbox}{max width=\linewidth}
\begin{tabular}{lrll}
\toprule
\textbf{Category} & Confirmed pool & \textbf{Ours} & \textbf{Measures} \\
\midrule
Repetition & 65 & 0.37 & $n$-gram 0.31, similarity 0.45 \\
Abandonment & 21 & 0.71 & switch marker 0.62 \\
After the answer & 38 & 0.87 & post-answer red.\ 0.13 \\
Infeasible Derivation & 27 & 1.00 & no measure \\
False Verdict & 8 & 1.00 & no measure \\
Unsupported deduction (R8--R11) + generic wrong step & 44 & 0.80 & PRM 0.09, confidence 0.11 \\
\bottomrule
\end{tabular}
\end{adjustbox}
\caption{\textbf{Pooled recall against the annotator-confirmed pool}, over the four annotators. Of the steps confirmed in a category (all carded strata pooled), the share each method's own flagged set contains. The pool is built from the carded strata, so this recall is relative to the cards. The last row pools the two unsupported-deduction categories with the confirmed steps of the generic wrong-step cards (9 steps, of which one carries an unsupported-deduction row of ours); it is the pool the unsupported-deduction rows of Table~\ref{tab:local-expert} are scored against, so that the pool is not defined by our cards alone.}
\label{tab:app-local-recall}
\end{table}

\paragraph{Precision and recall.}
Table~\ref{tab:app-local-precision} gives the full card counts behind the precision column of Table~\ref{tab:local-expert}, and Table~\ref{tab:app-local-recall} the pooled recall.
Over the five categories, the four annotators confirm 110 of the 176 decided cards of ours (0.63).
We traced every miss of ours on the first annotator's cards: three parser label losses (an illustration parsed as a derivation; two compound-action steps parsed as one action), three definitional calls by the annotator, five steps where the model is still finishing and the annotator read the doubt as part of that, and one abandonment with a true stated reason.

\paragraph{Restatement follow-up.}
The annotators' notion of repetition is wider than the tree's rows.
In a follow-up, we showed the first annotator 45 steps that restate earlier content (15 per task, each with the earlier step of the same content) under the standard repetition question.
The annotator confirmed 21 of 45 as repetition, split by what is restated: the puzzle's givens 5 of 20 (Game of 24 puzzle numbers 0 of 5, ZebraLogic clue recap 4 of 10, board echo 1 of 5) against the model's own earlier results 16 of 25 (Game of 24 unused-number list 4 of 6, expression 3 of 4, PlanBench own facts 9 of 15).
Counting restatements of the model's own results as repetition, our recall on that annotator's repetition cards moves from 0.36 to 0.43; about 5 of the 28 confirmed steps remain steps the engine did not match.

\paragraph{Lineage check on the undecided abandonment cards.}
On 11 abandonment cards, about 1/8 of the abandonment cards drawn (5 of ours, 6 switch-marker; 7 from the first annotator, 3 from the second, 1 from the fourth), the annotators could not decide from the text whether the branch being left was still workable.
When we showed them the engine's branch-viability check open, with the lineage of the node being left, they settled 9 of the 11 and agreed with the check on 7 of those (the two disagreements of the first annotator were on checks that found the branch viable, the one of the second on a check that found it exhausted); neither method's precision moved (ours 13/41 to 14/44, switch marker 10/38 to 12/44).
The check exists only for \arbor{}'s steps; for the switch-marker cards we took it from the same trajectory.

\subsection{What the local measures flag}
\begin{figure}[t]
    \centering
    \begin{subfigure}[b]{\linewidth}
        \centering
        \includegraphics[width=0.72\linewidth]{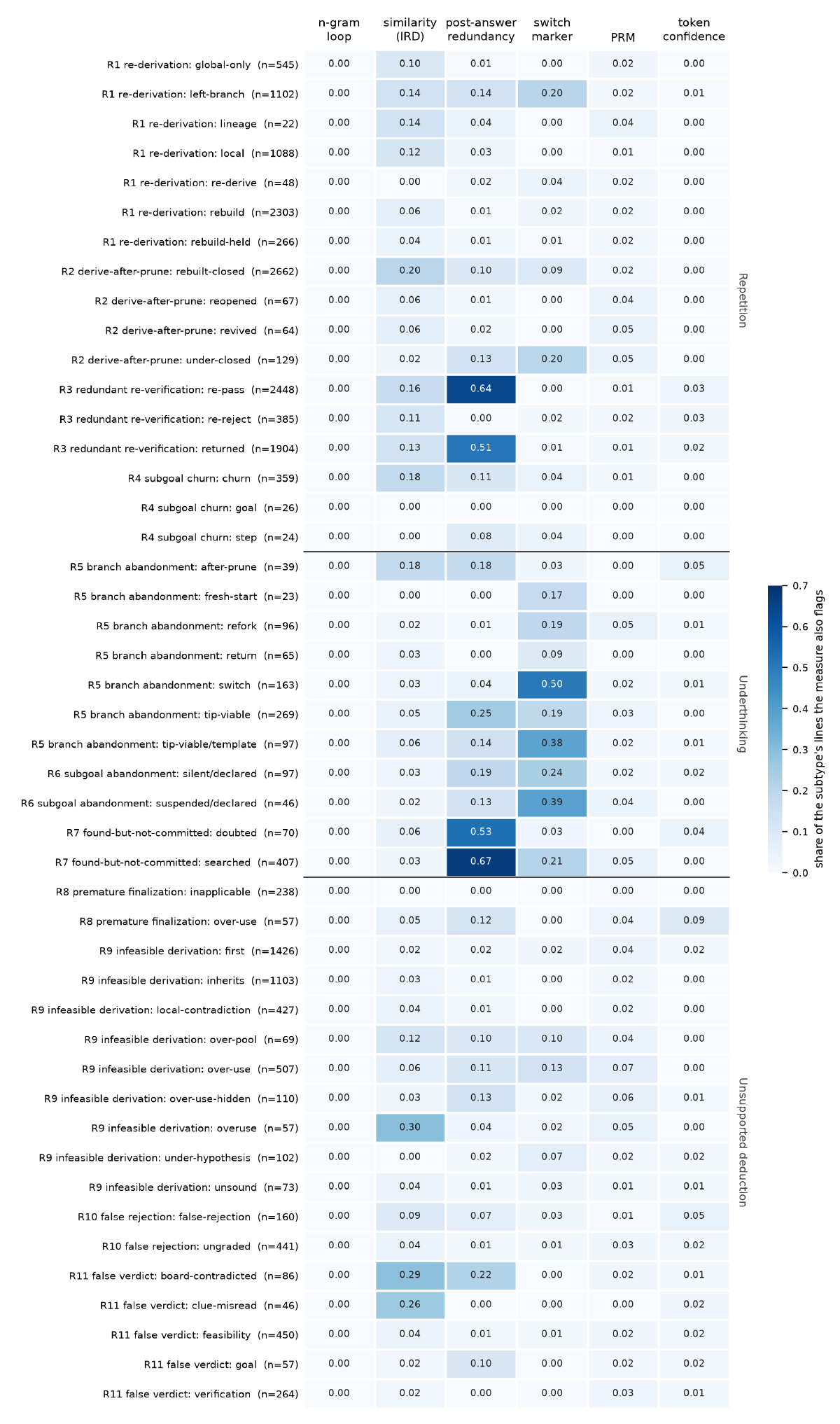}
        \caption{\textbf{Subtype coverage by each local measure} (\texttt{Qwen3-8B}, annotated sample of 342 traces per task). Each cell is the share of the subtype's steps that the measure also flags; $n$ is the subtype's step count on the sample. We group rows by family and number them as in Table~\ref{tab:error-rows}; the subtype name after the colon is the detector's own detail label. Post-answer redundancy reaches re-verification after the answer and post-answer attempt steps; the switch marker reaches the explicit ``switch'' form of branch abandonment; no measure reaches an unsupported-deduction subtype above 0.30, and the process reward model and token confidence stay at or below 0.09 on every subtype.}
        \label{fig:local-coverage}
    \end{subfigure}
    \vspace{4pt}
    \begin{subfigure}[b]{\linewidth}
        \centering
        \includegraphics[width=0.8\linewidth]{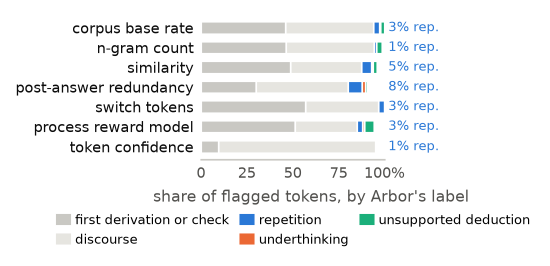}
        \caption{\textbf{Local signals flag ordinary reasoning, not failures.} What the tokens each local signal flags consist of, by \arbor{}'s labels (pooled over the three tasks, test questions), against the corpus base rate (top bar). Repetition shares are printed. We tune each signal to flag the same share of tokens as \arbor{}'s rows (\S\ref{sec:results-local-setup}); the published n-gram loop rule fires on 3 of 11{,}718 trajectories and is not shown. Per-task values in Appendix~\ref{sec:app-local}.}
        \label{fig:local-composition}
    \end{subfigure}
    \caption{What the local measures flag.}
    \label{fig:local-flag}
\end{figure}

\label{sec:app-local-switches}

\paragraph{Composition of the flagged sets.}
On the full corpus, most steps a local measure flags are first derivations or verifications (the forward work of the trajectory) and discourse (connective steps with no content, such as ``Wait'' or ``Let me see'').
With \arbor{}'s labels as the reference, the precision of each measure for its target category is near the base rate: similarity flags a repetition step with 5\% of its tokens, and a random step is one with 3\%.
The switches that switch tokens recover are mostly the explicit ``let me try another'' switches.

\begin{table}[t]
\centering
\footnotesize
\setlength{\tabcolsep}{4pt}
\begin{adjustbox}{max width=\linewidth}
\begin{tabular}{llrrr}
\toprule
\textbf{Switch set} & \textbf{Scope} & Yield (\%) & Non-viable & On R5/R6 step \\
\midrule
\multirow{4}{*}{Switch marker} & pooled & 3.9 & 0.50 {\scriptsize [0.49, 0.51]} & 0.033 \\
 & ZebraLogic & 0.9 & 0.58 {\scriptsize [0.55, 0.61]} & 0.051 \\
 & Game of 24 & 8.4 & 0.47 {\scriptsize [0.46, 0.48]} & 0.039 \\
 & PlanBench & 3.3 & 0.51 {\scriptsize [0.50, 0.52]} & 0.024 \\
\midrule
\multirow{4}{*}{Engine-detected} & pooled & 22.0 & 0.61 {\scriptsize [0.60, 0.62]} & 0.014 \\
 & ZebraLogic & 2.5 & 0.76 {\scriptsize [0.75, 0.77]} & 0.074 \\
 & Game of 24 & 14.1 & 0.71 {\scriptsize [0.70, 0.72]} & 0.065 \\
 & PlanBench & 32.1 & 0.60 {\scriptsize [0.59, 0.61]} & 0.005 \\
\bottomrule
\end{tabular}
\end{adjustbox}
\caption{\textbf{Switches by the marker and by the engine} (\texttt{Qwen3-8B}, test split). A marker switch is a step carrying a switch token; an engine-detected switch is a step on which the engine's attention cursor moves off a node. \emph{Yield}: share of tokens on switch steps. \emph{Non-viable}: share of switches that leave a branch the viability oracle marks non-viable at the switch step [95\% cluster-bootstrap interval over problems]. \emph{On R5/R6 step}: share of the set's steps that carry a branch or Subgoal Abandonment row. The marker reaches 0.25 of the abandonment steps pooled; the engine's own switch set, 0.67.}
\label{tab:app-local-switch}
\end{table}

\paragraph{Engine-detected switches against the switch marker.}
Table~\ref{tab:app-local-switch} compares the marker's switches with the switches the engine detects from the tree.
Engine-detected switches leave a branch the oracle marks non-viable more often than marker switches do; ``non-viable'' refers only to oracle viability at the switch step.
Few marker switches fall on an abandonment step.

\subsection{Outcome prediction per task and the length-normalized check}
\label{sec:app-local-outcome}
\begin{table*}[t]
\centering
\footnotesize
\setlength{\tabcolsep}{4pt}
\begin{adjustbox}{max width=\textwidth}
\begin{tabular}{l cccc cc}
\toprule
& \multicolumn{4}{c}{AUROC, incorrect vs.\ correct} & \multicolumn{2}{c}{Gain, pooled} \\
\cmidrule(lr){2-5} \cmidrule(lr){6-7}
\textbf{Score} & Pooled & Game of 24 & PlanBench & ZebraLogic$^{a}$ & $+$ rows & $+$ measure \\
\midrule
Length $+$ task (baseline) & 0.856 {\scriptsize [0.846, 0.864]} & 0.797 {\scriptsize [0.773, 0.818]} & 0.744 {\scriptsize [0.723, 0.764]} & 0.740 {\scriptsize [0.650, 0.818]} & & \\
\midrule
$n$-gram loop (count) & 0.880 {\scriptsize [0.872, 0.888]} & 0.796 {\scriptsize [0.772, 0.818]} & 0.773 {\scriptsize [0.752, 0.791]} & 0.740 {\scriptsize [0.651, 0.819]} & $+0.066$ & $+0.001$ \\
Similarity (IRD) & 0.866 {\scriptsize [0.857, 0.875]} & 0.818 {\scriptsize [0.796, 0.835]} & 0.875 {\scriptsize [0.863, 0.887]} & 0.739 {\scriptsize [0.653, 0.819]} & $+0.080$ & $+0.002$ \\
Post-answer redundancy (ERD)$^{b}$ & -- & -- & -- & -- & -- & -- \\
Switch marker (rate) & 0.893 {\scriptsize [0.885, 0.900]} & 0.801 {\scriptsize [0.776, 0.822]} & 0.825 {\scriptsize [0.809, 0.841]} & 0.737 {\scriptsize [0.652, 0.819]} & $+0.054$ & $+0.003$ \\
Process reward model (bottom-$k$) & 0.876 {\scriptsize [0.868, 0.885]} & 0.798 {\scriptsize [0.774, 0.820]} & 0.815 {\scriptsize [0.797, 0.831]} & 0.745 {\scriptsize [0.660, 0.821]} & $+0.072$ & $+0.004$ \\
Token confidence (bottom-$k$) & 0.873 {\scriptsize [0.865, 0.882]} & 0.799 {\scriptsize [0.774, 0.822]} & 0.777 {\scriptsize [0.759, 0.795]} & 0.747 {\scriptsize [0.665, 0.825]} & $+0.073$ & $+0.002$ \\
\midrule
Error-step count (all rows summed) & 0.863 {\scriptsize [0.855, 0.872]} & 0.796 {\scriptsize [0.774, 0.816]} & 0.743 {\scriptsize [0.722, 0.765]} & 0.739 {\scriptsize [0.652, 0.818]} & $+0.081$ & $-0.000$ \\
Per-row counts (ours) & \textbf{0.944} {\scriptsize [0.939, 0.950]} & \textbf{0.926} {\scriptsize [0.911, 0.941]} & \textbf{0.886} {\scriptsize [0.874, 0.898]} & 0.726 {\scriptsize [0.631, 0.803]} & & \\
\bottomrule
\end{tabular}
\end{adjustbox}
\caption{\textbf{What each score knows about the outcome} (\texttt{Qwen3-8B}, test split). AUROC of a logistic model for incorrect against correct traces, 5-fold cross-validation grouped by problem, with $\log$ trace length and the task as covariates in every model [95\% cluster-bootstrap interval over problems]; truncated traces excluded. \emph{$+$ rows}: gain in pooled AUROC from adding the per-row counts to the measure. \emph{$+$ measure}: gain from adding the measure to the per-row counts. The per-row vector is the count of steps carrying each row of Table~\ref{tab:error-rows}; the error-step count is their sum. $^{a}$ZebraLogic has 82 incorrect traces against 783 truncated, so incorrect-vs-correct is underpowered there; on truncated against completed, the per-row counts reach 0.986 [0.982, 0.990] (Appendix~\ref{sec:app-local}). $^{b}$Defined only on traces that reach a correct answer, which are all correct, so it has one outcome class.}
\label{tab:local-outcome}
\end{table*}

In Figure~\ref{fig:local-auroc} the five-signal ensemble closes about 3/5 of the gap between the length-and-task baseline and \arbor{}; length is a control because longer trajectories fail more often.
Because \arbor{}'s counts grow with trajectory length, we repeat the outcome prediction with every count normalized per 100 steps and no length term in the model: pooled, \arbor{} reaches 0.95 against 0.91 for the five local signals together and 0.84 for task alone, so the ordering of Figure~\ref{fig:local-auroc} is not an artifact of length (per-task values with intervals in Table~\ref{tab:local-outcome} and Table~\ref{tab:app-local-rates}). ZebraLogic has only 82 incorrect trajectories because its failures are truncations (783 of 877). There no signal, \arbor{} included, separates incorrect from correct beyond length; without the length term the local ensemble (0.75) edges \arbor{} (0.67). Truncation as the outcome is itself a matter of length on ZebraLogic (0.97 from length alone, 0.98 to 0.99 for every signal) but not pooled (0.96 for \arbor{} against 0.93 for the ensemble and 0.87 for length).
\begin{table}[t]
\centering
\scriptsize
\setlength{\tabcolsep}{3pt}
\begin{adjustbox}{max width=\linewidth}
\begin{tabular}{@{}lrrrr@{}}
\toprule
\textbf{Score} & \textbf{Zebra} & \textbf{24} & \textbf{Plan} & \textbf{Pooled} \\
\midrule
Task only (baseline) & 0.50 & 0.50 & 0.50 & 0.84 \\
n-gram & 0.67 & 0.76 & 0.51 & 0.86 \\
similarity (IRD) & 0.61 & 0.61 & 0.86 & 0.89 \\
switch tokens & 0.61 & 0.57 & 0.76 & 0.89 \\
process reward model & 0.69 & 0.65 & 0.72 & 0.89 \\
token confidence & 0.69 & 0.66 & 0.61 & 0.87 \\
all five local signals & 0.75 & 0.79 & 0.87 & 0.91 \\
\arbor{}: per-subtype rates & 0.67 & 0.92 & 0.87 & 0.95 \\
\bottomrule
\end{tabular}
\end{adjustbox}
\caption{Outcome AUROC (incorrect vs correct) with every count normalized by trajectory length and no length feature: each local signal as a rate, the five together, and \arbor{}'s per-subtype rates per 100 steps. Within a task the baseline is chance; pooled, the task indicator alone gives 0.84.}
\label{tab:app-local-rates}
\end{table}

Table~\ref{tab:app-local-rates} gives the length-free check per task. With every count divided by trajectory length and no length feature, \arbor{}'s rates lead the five local signals together on the 24~Game, tie on PlanBench and trail on ZebraLogic, where truncation is the dominant failure and length itself carries most of the signal.

\FloatBarrier
\section{Subtypes: definitions, full tables and robustness}
\label{sec:app-subtypes}
Rows undefined on a task (Table~\ref{tab:error-rows}): ZebraLogic posts no subgoals (R4, R6); a 24~Game check is an expression the engine evaluates, so a wrong verdict surfaces as R8 or R10 (R11); on PlanBench a step inside a closed branch also fails its preconditions, so we report its R2 as the subset of R9 that repeats a correctly pruned step. R8 firings on the emitted final answer record answer validity and are not step events.
Each subsection below is a table of the statistics defined for every subtype, followed by the statistics defined for one subtype only, stated in prose with their source; the main text reports selected values from these tables.
\begin{table*}[t]
\centering
\footnotesize
\setlength{\tabcolsep}{3pt}
\begin{adjustbox}{max width=\textwidth}
\begin{tabular}{l rrr r r r l l ccc}
\toprule
& \multicolumn{3}{c}{Rate per 100 steps} & Prev. & Tokens & Last & \multicolumn{2}{c}{Odds ratio, incorrect vs correct} & \multicolumn{3}{c}{Sign per task} \\
\cmidrule(lr){2-4} \cmidrule(lr){8-9} \cmidrule(lr){10-12}
Subtype & Zebra & 24 & Plan & (\%) & (\%) & third (\%) & Model A (between problems) & Model C (within problem) & Z & 24 & P \\
\midrule
\multicolumn{7}{l}{\textit{Repetition} (category)} & \textbf{0.45 {\scriptsize [0.34, 0.61]}} & 0.83 {\scriptsize [0.53, 1.32]} & & & \\
\quad R1 Re-Derivation & 2.65 & 1.08 & 0.12 & 59 & 1.0 & 30 & 0.98 {\scriptsize [0.86, 1.13]} & \textbf{1.49 {\scriptsize [1.18, 1.88]}} & $\triangledown$ & $\blacktriangle$ & $\circ$ \\
\quad R2 Derive-after-Prune & 0.34 & 2.62 & 0.18 & 56 & 1.0 & 40 & \textbf{0.66 {\scriptsize [0.58, 0.76]}} & 0.90 {\scriptsize [0.72, 1.14]} & $\circ$ & $\circ$ & $\triangledown$ \\
\quad R3 Redundant Re-Verification & 0.78 & 3.11 & 0.59 & 82 & 1.2 & 60 & \textbf{0.37 {\scriptsize [0.31, 0.44]}} & \textbf{0.48 {\scriptsize [0.37, 0.63]}} & $\triangledown$ & $\triangledown$ & $\triangledown$ \\
\quad R4 Subgoal Churn & -- & 0.35 & 0.04 & 29 & 0.2 & 25 & \textbf{1.28 {\scriptsize [1.12, 1.51]}} & 1.24 {\scriptsize [0.98, 1.57]} & -- & $\blacktriangle$ & $\circ$ \\
\midrule
\multicolumn{7}{l}{\textit{Underthinking} (category)} & \textbf{0.57 {\scriptsize [0.50, 0.66]}} & \textbf{0.42 {\scriptsize [0.33, 0.53]}} & & & \\
\quad R5 Branch Abandonment & 0.20 & 0.42 & 0.13 & 47 & 0.3 & 25 & 1.07 {\scriptsize [0.95, 1.20]} & 0.83 {\scriptsize [0.69, 1.02]} & $\circ$ & $\triangledown$ & $\blacktriangle$ \\
\quad R6 Subgoal Abandonment & -- & 0.14 & 0.009 & 16 & 0.06 & 32 & \textbf{1.24 {\scriptsize [1.03, 1.46]}} & 1.18 {\scriptsize [0.89, 1.56]} & -- & $\circ$ & $\circ$ \\
\quad R7 Post-Answer Attempt & 0.03 & 0.53 & 0.03 & 24 & 0.2 & 47 & \textbf{0.08 {\scriptsize [0.07, 0.10]}} & \textbf{0.14 {\scriptsize [0.10, 0.19]}} & $\circ$ & $\triangledown$ & $\triangledown$ \\
\midrule
\multicolumn{7}{l}{\textit{Unsupported deduction} (category)} & \textbf{4.68 {\scriptsize [3.30, 7.29]}} & \textbf{8.34 {\scriptsize [4.98, 17.68]}} & & & \\
\quad R8 Premature Finalization & 0.008 & 0.08 & 0.003 & 7 & 0.04 & 70 & \textbf{10.12 {\scriptsize [8.15, 12.85]}} & \textbf{10.25 {\scriptsize [6.98, 16.70]}} & $\blacktriangle$ & $\blacktriangle$ & $\blacktriangle$ \\
\quad\quad mid-trace commits only & & & & & & & \textbf{2.11 {\scriptsize [1.59, 2.80]}} & 1.41 {\scriptsize [0.92, 2.26]} & & & \\
\quad R9 Infeasible Derivation & 1.12 & 0.85 & 0.89 & 74 & 1.2 & 38 & \textbf{2.90 {\scriptsize [2.35, 3.60]}} & \textbf{4.57 {\scriptsize [3.22, 7.03]}} & $\circ$ & $\blacktriangle$ & $\blacktriangle$ \\
\quad R10 False Rejection & 0.08 & 0.05 & 0.27 & 30 & 0.2 & 39 & \textbf{0.87 {\scriptsize [0.76, 0.99]}} & 1.03 {\scriptsize [0.84, 1.27]} & $\circ$ & $\circ$ & $\triangledown$ \\
\quad R11 False Verdict & 0.13 & -- & 0.50 & 48 & 0.4 & 52 & \textbf{1.70 {\scriptsize [1.46, 1.96]}} & 1.19 {\scriptsize [0.93, 1.56]} & $\blacktriangle$ & -- & $\blacktriangle$ \\
\bottomrule
\end{tabular}
\end{adjustbox}
\caption{\textbf{Each subtype on its own} (Qwen3-8B, 11{,}718 traces; the 40 engine-development problems per task excluded). Rate: events per 100 steps per task. Prev.: share of traces with at least one event, over the tasks that define the subtype. Tokens: share of all tokens on steps carrying the subtype (Appendix~\ref{sec:app-token-share}). Last third: share of the subtype's events in the last third of the trace. Odds ratios of the subtype's presence with 95\% cluster-bootstrap intervals over problems: Model A, incorrect vs correct at equal log length and task, truncated traces excluded; Model C, the same within the 785 problems that have both outcomes (conditional logit). Category rows give the odds ratio of the category's presence. Bold: Benjamini--Hochberg $q<0.05$. Sign per task (Model A): $\blacktriangle$ above 1, $\triangledown$ below 1, $\circ$ not significant, -- undefined (Table~\ref{tab:error-rows}). Premature Finalization (R8) fires on a finalization that fails the acceptance test, so its odds ratio is largely its definition; the sub-row counts only mid-trace commits the model later revised. PlanBench R2 is the subset of its R9 steps that repeat a correctly pruned step.}
\label{tab:subtypes}
\end{table*}

\subsection{Tree shapes}
\label{sec:app-tree-shapes}
Figure~\ref{fig:tree-shapes} (\S\ref{sec:results-subtypes-setup}) gives the schematic tree of a typical failing trajectory per task. A ZebraLogic trajectory is one deep chain of consecutive derivations (a maximum depth of 23 for the median trajectory) that switches lines of reasoning only about every 40 steps, split at a few case assumptions whose rejected cases are pruned; a 24~Game trajectory is a shallow fan (maximum depth 7) whose nodes are mostly siblings, alternative compositions tried in turn from the same starting point, with a switch between them every 7 steps; a PlanBench trajectory is a single chain of moderate depth (maximum depth 16) that appends one plan step after another and rarely branches, but 1/4 of its nodes are states the model assumes rather than derives, and what it checks are the preconditions each step needs.
\subsection{Token share and per-task cost}
\label{sec:app-token-share}
The token share of a subtype is the share of all tokens of a task's traces that lie on steps carrying the subtype, under the Qwen3 tokenizer; a step carrying two subtypes counts for both. It measures the cost of the steps a detector labels. Table~\ref{tab:app-token-share} gives the per-task values behind the pooled column of Table~\ref{tab:subtypes}.
\begin{table}[t]
\centering
\footnotesize
\setlength{\tabcolsep}{3pt}
\begin{adjustbox}{max width=\linewidth}
\begin{tabular}{l rrrr rrrr}
\toprule
& \multicolumn{4}{c}{Token share (\%)} & \multicolumn{4}{c}{Prevalence (\%)} \\
\cmidrule(lr){2-5} \cmidrule(lr){6-9}
Subtype & Zebra & 24 & Plan & Pooled & Zebra & 24 & Plan & Pooled \\
\midrule
R1 Re-Derivation & 3.17 & 1.37 & 0.15 & 1.05 & 87 & 69 & 35 & 59 \\
R2 Derive-after-Prune & 0.48 & 3.58 & 0.23 & 0.98 & 48 & 82 & 41 & 56 \\
R3 Redundant Re-Verification & 0.72 & 3.33 & 0.56 & 1.18 & 78 & 92 & 76 & 82 \\
R4 Subgoal Churn & -- & 0.45 & 0.05 & 0.16 & -- & 45 & 15 & 29 \\
R5 Branch Abandonment & 0.26 & 0.56 & 0.16 & 0.27 & 44 & 60 & 38 & 47 \\
R6 Subgoal Abandonment & -- & 0.20 & 0.01 & 0.06 & -- & 31 & 4 & 16 \\
R7 Post-Answer Attempt & 0.03 & 0.85 & 0.03 & 0.20 & 8 & 57 & 7 & 24 \\
R8 Premature Finalization & 0.01 & 0.07 & 0.04 & 0.04 & 3 & 18 & 2 & 7 \\
R9 Infeasible Derivation & 1.32 & 1.50 & 1.09 & 1.23 & 60 & 56 & 96 & 74 \\
R10 False Rejection & 0.07 & 0.03 & 0.28 & 0.18 & 20 & 11 & 52 & 30 \\
R11 False Verdict & 0.11 & -- & 0.58 & 0.45 & 14 & -- & 68 & 48 \\
\midrule
Any subtype (a step counted once) & 6.1 & 11.7 & 3.4 & 5.7 & 93 & 100 & 100 & 98 \\
\bottomrule
\end{tabular}
\end{adjustbox}
\caption{Token share and prevalence per task (Qwen3-8B). Token share: tokens on steps carrying the subtype over all tokens of the task's traces (Qwen3 tokenizer); a step carrying two subtypes counts for both. Pooled prevalence is over the tasks that define the subtype. The last row counts every detector row, including rows the paper does not report.}
\label{tab:app-token-share}
\end{table}

\subsection{Extended cost: the tokens an event opens}
\label{sec:app-extended-cost}
\begin{table*}[t]
\centering
\footnotesize
\setlength{\tabcolsep}{4pt}
\begin{tabular}{@{}l rrrr rrr@{}}
\toprule
& \multicolumn{4}{c}{Median tokens opened / share of the trajectory (\%)} & \multicolumn{3}{c}{Pooled share by outcome (\%)} \\
\cmidrule(lr){2-5} \cmidrule(lr){6-8}
Subtype & ZebraLogic & 24 Game & PlanBench & Pooled & correct & incorrect & truncated \\
\midrule
\multicolumn{8}{@{}l}{\textit{Repetition}} \\
R1 Re-Derivation & 17 / 0.3 & 15 / 0.4 & 16 / 0.2 & 16 / 0.3 & 0.4 & 0.3 & 0.2 \\
R2 Derive-after-Prune & 25 / 0.4 & 17 / 0.4 & 220 / 2.7 & 20 / 0.4 & 0.5 & 0.5 & 0.3 \\
R3 Redundant Re-Verification & 11 / 0.2 & 24 / 0.8 & 12 / 0.2 & 16 / 0.3 & 0.6 & 0.2 & 0.2 \\
R4 Subgoal Churn & -- & 14 / 0.3 & 17 / 0.2 & 15 / 0.3 & 0.4 & 0.2 & 0.2 \\
\midrule
\multicolumn{8}{@{}l}{\textit{Underthinking}} \\
R5 Branch Abandonment & 948 / 22.8 & 430 / 12.5 & 2,886 / 37.6 & 906 / 20.8 & 14.6 & 33.0 & 15.0 \\
R6 Subgoal Abandonment & -- & 1,381 / 47.8 & 2,825 / 39.8 & 1,489 / 46.7 & 50.6 & 39.9 & 43.0 \\
R7 Post-Answer Attempt & 267 / 10.5 & 18 / 0.6 & 256 / 5.3 & 19 / 0.7 & 0.7 & 0.7 & 0.4 \\
\midrule
\multicolumn{8}{@{}l}{\textit{Unsupported deduction}} \\
R8 Premature Finalization & 446 / 9.5 & 201 / 4.9 & 49 / 0.6 & 208 / 4.9 & 6.1 & 4.2 & 4.6 \\
R9 Infeasible Derivation & 213 / 3.1 & 57 / 1.3 & 229 / 2.9 & 173 / 2.6 & 2.1 & 2.9 & 2.2 \\
R10 False Rejection & 1,947 / 34.7 & 1,691 / 45.4 & 3,070 / 37.1 & 2,828 / 37.5 & 42.9 & 36.4 & 31.8 \\
R11 False Verdict & 340 / 5.9 & -- & 25 / 0.3 & 30 / 0.4 & 0.5 & 0.3 & 3.0 \\
\bottomrule
\end{tabular}
\caption{\textbf{Extended cost: the tokens an event opens} (Qwen3-8B). For each event, the region it opens is, for Branch and Subgoal Abandonment, the steps from the event until the model returns to the left branch, else to the end; for Post-Answer Attempt, until the commit; for Premature Finalization, Infeasible Derivation and False Verdict, every later step resolved to the event's node or a descendant; for Derive-after-Prune, the later steps in the pruned subtree; for Re-Derivation, Redundant Re-Verification and Subgoal Churn, the steps of the burst the event starts; and for False Rejection, the steps until the rejected subtree is next entered, else to the end. --: undefined on the task.}
\label{tab:app-extended-cost}
\end{table*}

The token share counts the labeled steps only; the extended cost counts the region an event opens, with one row per event and one per burst for the repetition rows (Table~\ref{tab:app-extended-cost}).
Regions of different events overlap and we do not sum them within a trajectory.
The budget a single event opens is large only for the underthinking rows and for False Rejection, whose region is the work done away from a viable branch, and the model re-enters the rejected branch in 5\% of cases pooled; a Premature Finalization's subtree reaches the committed lineage with 45\% of its tokens, and an Infeasible Derivation's with 10\% pooled and 33\% on the 24~Game.
The region covers what follows the event, without causal attribution; a span to the end of the trajectory scales with where the event falls, and burst tokens count event steps only.
\subsection{Full odds-ratio tables}
\label{sec:app-odds-ratios}
Table~\ref{tab:app-or-pooled} gives every model on the pooled data. Model B (truncated vs completed) separates budget from accuracy: Derive-after-Prune (1.42) and Infeasible Derivation (1.83) go with running out of budget, Redundant Re-Verification (0.45) and Branch Abandonment (0.64) with completing, and Premature Finalization is not associated with truncation (0.78, n.s.) although it has the largest incorrect-answer odds ratio.
\begin{table*}[t]
\centering
\footnotesize
\setlength{\tabcolsep}{4pt}
\begin{adjustbox}{max width=\linewidth}
\begin{tabular}{lllll}
\toprule
Subtype & A: incorrect vs correct & A$'$: rate instead of presence & B: truncated vs completed & C: within problem \\
\midrule
\textit{Repetition} (category) & \textbf{0.45 {\scriptsize [0.34, 0.61]}} & \textbf{0.79 {\scriptsize [0.75, 0.82]}} & \textbf{0.15 {\scriptsize [0.06, 0.60]}} & 0.83 {\scriptsize [0.53, 1.32]} \\
\quad R1 Re-Derivation & 0.98 {\scriptsize [0.86, 1.13]} & \textbf{1.18 {\scriptsize [1.05, 1.30]}} & 0.82 {\scriptsize [0.46, 1.52]} & \textbf{1.49 {\scriptsize [1.18, 1.88]}} \\
\quad R2 Derive-after-Prune & \textbf{0.66 {\scriptsize [0.58, 0.76]}} & \textbf{1.19 {\scriptsize [1.11, 1.28]}} & \textbf{1.42 {\scriptsize [1.11, 1.83]}} & 0.90 {\scriptsize [0.72, 1.14]} \\
\quad R3 Redundant Re-Verification & \textbf{0.37 {\scriptsize [0.31, 0.44]}} & \textbf{0.70 {\scriptsize [0.67, 0.74]}} & \textbf{0.45 {\scriptsize [0.33, 0.60]}} & \textbf{0.48 {\scriptsize [0.37, 0.63]}} \\
\quad R4 Subgoal Churn & \textbf{1.28 {\scriptsize [1.12, 1.51]}} & \textbf{1.92 {\scriptsize [1.56, 2.39]}} & 1.15 {\scriptsize [0.84, 1.66]} & 1.24 {\scriptsize [0.98, 1.57]} \\
\midrule
\textit{Underthinking} (category) & \textbf{0.57 {\scriptsize [0.50, 0.66]}} & \textbf{0.39 {\scriptsize [0.33, 0.45]}} & \textbf{0.54 {\scriptsize [0.44, 0.67]}} & \textbf{0.42 {\scriptsize [0.33, 0.53]}} \\
\quad R5 Branch Abandonment & 1.07 {\scriptsize [0.95, 1.20]} & 0.91 {\scriptsize [0.77, 1.07]} & \textbf{0.64 {\scriptsize [0.52, 0.78]}} & 0.83 {\scriptsize [0.69, 1.02]} \\
\quad R6 Subgoal Abandonment & \textbf{1.24 {\scriptsize [1.03, 1.46]}} & \textbf{1.58 {\scriptsize [1.07, 2.18]}} & 1.18 {\scriptsize [0.87, 1.62]} & 1.18 {\scriptsize [0.89, 1.56]} \\
\quad R7 Post-Answer Attempt & \textbf{0.08 {\scriptsize [0.07, 0.10]}} & \textbf{0.08 {\scriptsize [0.04, 0.12]}} & 0.77 {\scriptsize [0.53, 1.15]} & \textbf{0.14 {\scriptsize [0.10, 0.19]}} \\
\midrule
\textit{Unsupported deduction} (category) & \textbf{4.68 {\scriptsize [3.30, 7.29]}} & \textbf{1.61 {\scriptsize [1.50, 1.73]}} & \textbf{2.66 {\scriptsize [1.60, 5.66]}} & \textbf{8.34 {\scriptsize [4.98, 17.68]}} \\
\quad R8 Premature Finalization & \textbf{10.12 {\scriptsize [8.15, 12.85]}} & \textbf{154.13 {\scriptsize [96.90, 272.89]}} & 0.78 {\scriptsize [0.54, 1.09]} & \textbf{10.25 {\scriptsize [6.98, 16.70]}} \\
\quad R9 Infeasible Derivation & \textbf{2.90 {\scriptsize [2.35, 3.60]}} & \textbf{2.18 {\scriptsize [2.01, 2.37]}} & \textbf{1.83 {\scriptsize [1.22, 3.03]}} & \textbf{4.57 {\scriptsize [3.22, 7.03]}} \\
\quad R10 False Rejection & \textbf{0.87 {\scriptsize [0.76, 0.99]}} & 0.82 {\scriptsize [0.66, 1.02]} & 1.00 {\scriptsize [0.80, 1.23]} & 1.03 {\scriptsize [0.84, 1.27]} \\
\quad R11 False Verdict & \textbf{1.70 {\scriptsize [1.46, 1.96]}} & \textbf{2.26 {\scriptsize [1.97, 2.70]}} & 0.91 {\scriptsize [0.66, 1.25]} & 1.19 {\scriptsize [0.93, 1.56]} \\
\bottomrule
\end{tabular}
\end{adjustbox}
\caption{Odds ratios on the three-task pool (Qwen3-8B), 95\% cluster-bootstrap intervals over problems. A: presence of the subtype, incorrect vs correct, truncated excluded, with log length and task as covariates. A$'$: the subtype's rate per 100 steps (standardized) in place of presence. B: truncated vs completed. C: conditional logit within the problems that have both outcomes (785 strata; 739 for R4 and R6, 474 for R11). Bold: Benjamini--Hochberg $q<0.05$ within the column. The category rows include every detector row of the category, as computed.}
\label{tab:app-or-pooled}
\end{table*}

\subsection{The event and the committed answer}
\label{sec:app-answer-path}
\begin{table}[t]
\centering
\footnotesize
\setlength{\tabcolsep}{2.5pt}
\begin{adjustbox}{max width=\linewidth}
\begin{tabular}{@{}l rrr rrr@{}}
\toprule
& \multicolumn{3}{c}{Events on the answer's path (\%)} & \multicolumn{3}{c}{Answers whose path carries the row (\%)} \\
& & & & \multicolumn{3}{c}{wrong / correct} \\
\cmidrule(lr){2-4} \cmidrule(lr){5-7}
Subtype & Zebra & 24 & Plan & Zebra & 24 & Plan \\
\midrule
R1 Re-Derivation & 17 & 4 & 15 & 87 / 69 & 19 / 9 & 11 / 10 \\
R2 Derive-after-Prune & 4 & 1 & 15 & 15 / 6 & 8 / 4 & 26 / 3 \\
R3 Redundant Re-Verification & 46 & 67 & 28 & 52 / 60 & 70 / 93 & 43 / 55 \\
R4 Subgoal Churn & -- & 3 & 4 & -- & 5 / 2 & 1 / 1 \\
R5 Branch Abandonment & 23 & 4 & 7 & 23 / 20 & 3 / 5 & 6 / 8 \\
R6 Subgoal Abandonment & -- & 0 & 0 & -- & -- & -- \\
R7 Post-Answer Attempt & 44 & 67 & 30 & 6 / 6 & 6 / 46 & -- / 14 \\
R8 Premature Finalization & 96 & 43 & 75 & 24 / 4 & 60 / 1 & 2 / 0 \\
R9 Infeasible Derivation & 6 & 14 & 9 & 35 / 15 & 65 / 1 & 79 / 13 \\
R10 False Rejection & 11 & 5 & 1 & 10 / 4 & 0 / 1 & 2 / 2 \\
R11 False Verdict & 37 & -- & 11 & 23 / 3 & -- & 21 / 10 \\
\bottomrule
\end{tabular}
\end{adjustbox}
\caption{\textbf{The event and the committed answer} (Qwen3-8B). Left: the share of the subtype's events whose node lies on the dependency path of the trajectory's committed answer. Right: the share of incorrect, and of correct, committed answers whose dependency path contains at least one event of the subtype (the answer-path share of \S\ref{sec:results-subtypes-setup}); we exclude truncated trajectories, 19\% of which on ZebraLogic commit a step they later leave. --: undefined on the task, or no committed answers of that outcome carry the row.}
\label{tab:app-answer}
\end{table}

Table~\ref{tab:app-answer} gives, per subtype, how its events relate to the committed answer.
Beyond the Infeasible Derivation, correct answers' paths carry little but Redundant Re-Verification.
The wrong answers without an Infeasible Derivation on their path, 1/4 of the total, fail elsewhere: on the 24~Game 1/3 of them commit an answer against the model's own state (Premature Finalization) and on PlanBench 1/4 rest on a False Verdict, while in 1/3 of PlanBench's and 1/5 of the 24~Game's the path carries none of the failure rows we define.
Reading those cases shows two other causes: a plan that stops short of the goal (1/6 of PlanBench's), and a misread clue or initial state, which makes a sound-looking path wrong and is present in about 4/5 of these trajectories, though as often in other wrong ones.
\paragraph{Premature Finalization.}
Of the 915 premature commits on the 24~Game, 66\% rest on a lineage containing an Infeasible Derivation, 43\% one fired within the previous ten steps, 78\% on a lineage the oracle marks non-viable and 75\% on a lineage with an unverified node; 67\% had passed a check on the committed node itself; 9\% rest on nothing a row names, mostly commits of an unattached root.
On ZebraLogic 44\% of 92 premature commits and on PlanBench 38\% of 76 rest on a lineage containing an Infeasible Derivation.
When the model later revises such a commit, it does so by committing again, and a failed check precedes the revision in only 7\% of cases.
\paragraph{Derive-after-Prune.}
When the prune was right, the rebuilt node reaches the committed answer in 1.1\% of events on ZebraLogic, 0.5\% on the 24~Game and 14.9\% on PlanBench; when it was wrong, an implicit repair of a False Rejection, in 33.4\% on ZebraLogic and 2.0\% on the 24~Game.
\paragraph{Branch Abandonment.}
On the 24~Game the abandoned subtree holds the trajectory's own solution node in 6.9\% of events.
\paragraph{False Verdict.}
In non-truncated ZebraLogic trajectories a false pass sits on the committed answer's lineage in 58.6\% of events; a false fail lands on the solution node or an ancestor in 3.8\% of ZebraLogic events and 2.1\% of PlanBench events.

\subsection{The state at the event}
\label{sec:app-state}
\begin{table}[t]
\centering
\footnotesize
\setlength{\tabcolsep}{3pt}
\begin{adjustbox}{max width=\linewidth}
\begin{tabular}{@{}ll rrr@{}}
\toprule
Subtype & Earlier state referred to & Zebra & 24 & Plan \\
\midrule
\multicolumn{5}{@{}l}{\textit{A. Median steps from the event back to that state}} \\
R1 Re-Derivation & the result re-derived & 29 & 70 & 55 \\
R2 Derive-after-Prune & the prune of the branch & 12 & 79 & 73 \\
R3 Redundant Re-Verification & the check repeated & 21 & 15 & 39 \\
R4 Subgoal Churn & last entry on the subgoal$^{a}$ & $\times$ & 82 & -- \\
R5 Branch Abandonment & the branch's last new node$^{b}$ & -- & 13 & -- \\
\midrule
\multicolumn{5}{@{}l}{\textit{B. Share of events where the model announces the change (\%)}} \\
R5 Branch Abandonment & explicit prune on the step & -- & 9.5$^{c}$ & $\circ$ \\
R2 Derive-after-Prune & explicit re-open before deriving & 15.8 & 3.5 & $\circ$ \\
\bottomrule
\end{tabular}
\end{adjustbox}
\caption{\textbf{The state at the event.} A: how far back the earlier state lies that the event repeats, re-enters or leaves, as the median number of steps over the subtype's events (Qwen3-8B). B: how often the model states the change in text; the remainder are silent. $\times$: row undefined on the task; --: defined but not yet computed; $\circ$: 0 by construction, PlanBench has no branch lifecycle. $^{a}$Against 4 steps for subgoals the model completes. $^{b}$Against 20 on the steps where an abandonment could have occurred. $^{c}$A second record on the same events gives 11.2\%; we are reconciling the two definitions.}
\label{tab:app-state}
\end{table}

Table~\ref{tab:app-state} gives, per subtype, how far back the earlier state lies that the event refers to, and whether the model announces the change.
The facts that do not generalize across subtypes are these.
A Re-Derivation's copy sits on the source's own lineage in 60\% of events on ZebraLogic, 24\% on the 24~Game and 86\% on PlanBench, and across a switch of the source's branch in 32\%, 50\% and 14\%; on the 24~Game it is a median 12 switches and 7 prunes from its source; the copy is a new node in 38\%, 8\% and 11\%.
A Branch Abandonment leaves from a frontier leaf, a node with no child yet, in 67\% of events on ZebraLogic, 53\% on the 24~Game and 41\% on PlanBench; where the model goes next is a task property, a restart from the root on the 24~Game (66\%), a cousin case on ZebraLogic (60\%) and a sibling step on PlanBench (40\%).
A Subgoal Churn on the 24~Game leaves a subgoal that had gained a node since its last entry in 63\% of events, against 83\% for subgoals the model completes.
A Derive-after-Prune re-enters a branch whose pruned node was the model's last open option in 47\% of events on the 24~Game and 27\% on ZebraLogic.

\subsection{What follows an event}
\label{sec:app-follows}
\label{sec:app-departures}
\begin{table*}[t]
\centering
\footnotesize
\setlength{\tabcolsep}{4pt}
\begin{adjustbox}{max width=\textwidth}
\begin{tabular}{@{}l l rrr@{}}
\toprule
Subtype & What follows the event & Zebra & 24 & Plan \\
\midrule
\multicolumn{5}{@{}l}{\textit{Unsupported deduction}} \\
R9 Infeasible Derivation & never checked / caught by the first later check / passed by it & 59.1 / 4.7 / 36.2 & 44.2 / 24.1 / 31.7 & 40.0 / 30.8 / 29.1$^{a}$ \\
R8 Premature Finalization & stands as the final commit / revised later & \multicolumn{3}{c}{50.6 / 49.4 (pooled)} \\
R11 False Verdict & false pass / false fail & 92.5 / 7.5 & -- & 34.4 / 65.6 \\
R10 False Rejection & rejected branch re-entered / never & 21 / 79 & 18 / 82 & 1.5 / 98.5 \\
\midrule
\multicolumn{5}{@{}l}{\textit{Underthinking}} \\
R5 Branch Abandonment & node closed before the event / on it / later / never & 6.2 / 0.0 / 10.1 / 83.7 & 6.6 / 7.9 / 3.2 / 82.3 & $\circ$ \\
 & model returns to the branch / never & 42.9 / 57.1 & 70.7 / 29.3 & 22.0 / 78.0 \\
R6 Subgoal Abandonment & subgoal resumed / never & -- & \multicolumn{2}{c}{6 / 94 (pooled)} \\
R7 Post-Answer Attempt$^{b}$ & alternative branch opened / not & 52.2 / 47.8 & 89.5 / 10.5 & 97.0 / 3.0 \\
 & ends: wrong commit / truncated / commits the solution / no commit & 13.4 / 67.2 / 14.9 / 4.5 & 54.3 / 34.1 / 9.7 / 1.9 & 77.2 / 0.7 / 0.4 / 21.7 \\
\midrule
\multicolumn{5}{@{}l}{\textit{Repetition}} \\
R2 Derive-after-Prune & re-entered branch was dead / wrongly closed & 90.6 / 9.4 & 78.1 / 21.9 & 100 / 0$^{c}$ \\
 & re-entered silently / with an explicit re-open & 84.2 / 15.8 & 96.5 / 3.5 & 100 / 0$^{c}$ \\
R1 Re-Derivation & later work builds on the original / on the copy / neither & 52 / 19 / 24 & 67 / 3 / 27 & -- \\
R4 Subgoal Churn & model returns to the subgoal / never & -- & 89 / 11 & -- \\
R3 Redundant Re-Verification & verdict unchanged / flipped & 99.5 / 0.5 & 99.2 / 0.8 & 95.1 / 4.9 \\
\bottomrule
\end{tabular}
\end{adjustbox}
\caption{\textbf{What follows an event.} For each subtype, one exhaustive partition of what the trajectory does next, read from the engine's state, in percent of the subtype's events (Qwen3-8B; rows sum to 100 within rounding). --: undefined or not measured; $\circ$: PlanBench has no branch lifecycle. $^{a}$0.1\% mixed. $^{b}$Over the failing trajectories that held a correct solution. $^{c}$By construction: on PlanBench the row is defined as a repeat of a correctly pruned step and there is no re-open operation.}
\label{tab:app-follows}
\end{table*}

Table~\ref{tab:app-follows} gives, per subtype, one exhaustive partition of what the trajectory does next.
Behind the Infeasible Derivation row: among the events that are verified, the verification passes the step in 57\% of cases on the 24~Game, 89\% on ZebraLogic and 49\% on PlanBench, and a False Verdict is later flagged on the same node in 3.1\% of ZebraLogic and 7.3\% of PlanBench false passes.
Behind the Branch Abandonment rows: a return closes the branch within ten steps in 3\% of events on the 24~Game and 2\% on ZebraLogic, and the returning step is a derivation in 80\% and 64\%.
Behind the Post-Answer Attempt rows: a standing correct solution is falsely failed in under 1\% of cases, and trajectories that open no new branch after it commit within a dozen steps; Table~\ref{tab:app-fca} gives the failing trajectories that held a correct solution, which on the 24~Game and PlanBench spend more of their tokens after that solution than a correct trajectory does (64 against 57\% and 51 against 33\%) and on ZebraLogic less (18 against 24\%).
Behind the Redundant Re-Verification row: a flipped verdict is followed by a prune within ten steps in 71\% of cases on ZebraLogic, 47\% on the 24~Game and 38\% on PlanBench, against 4\%, 9\% and 16\% after an unchanged one.
\begin{table}[t]
\centering
\footnotesize
\setlength{\tabcolsep}{3pt}
\begin{adjustbox}{max width=\linewidth}
\begin{tabular}{l llll}
\toprule
& Zebra & 24~Game & Plan & Pooled \\
\midrule
Failing traces with a correct solution (\%) & 17.3 {\scriptsize [14.4, 20.1]} & 30.8 {\scriptsize [27.2, 34.4]} & 14.8 {\scriptsize [13.4, 16.2]} & 17.7 {\scriptsize [16.6, 18.9]} \\
Token share after it, failing (\%) & 18 & 64 & 51 & 48 \\
Token share after it, correct (\%) & 24 & 57 & 33 & 40 \\
\midrule
\multicolumn{5}{l}{Share of those failing traces with the subtype at or after the solution (\%)} \\
R1 Re-Derivation & 34 & 62 & 30 & 40 \\
R2 Derive-after-Prune & 12 & 78 & 41 & 47 \\
R3 Redundant Re-Verification & 48 & 82 & 72 & 71 \\
R4 Subgoal Churn & -- & 39 & 10 & 20 \\
R5 Branch Abandonment & 13 & 63 & 30 & 36 \\
R6 Subgoal Abandonment & -- & 22 & 2 & 9 \\
R7 Post-Answer Attempt & 12 & 62 & 10 & 24 \\
R8 Premature Finalization & 3 & 45 & 1 & 13 \\
R9 Infeasible Derivation & 15 & 69 & 81 & 67 \\
R10 False Rejection & 6 & 13 & 40 & 27 \\
R11 False Verdict & 7 & -- & 56 & 45 \\
\bottomrule
\end{tabular}
\end{adjustbox}
\caption{Failing traces that held a correct complete solution at some step (Qwen3-8B). ``Failing'' is incorrect or truncated. The first step of a correct complete solution on a node of the tree is the reference point; intervals are 95\% cluster bootstraps over problems. The population is a count, not a gain.}
\label{tab:app-fca}
\end{table}

\subsection{Where the budget goes}
\label{sec:app-budget}
Three estimates locate the budget; all three read the 8B step table (one row per step with its token count, the node it lands on, \arbor{}'s rows, the node's oracle viability, and whether the node lies on the committed answer's dependency path).
\emph{Off-path share}: the share of a trajectory's tokens on steps whose node is not on the final answer's dependency path; undefined for truncated trajectories, which commit nothing.
\emph{Dead-end share}: the share of tokens on steps whose node the viability oracle marks as unable to reach a correct solution and that carry no failure row. Within questions that have both outcomes, the median truncated trajectory spends about twice the dead-end tokens of its completed sibling (ZebraLogic 5{,}065 vs 2{,}588; 24~Game 3{,}180 vs 1{,}580; PlanBench 10{,}860 vs 4{,}724) while its failure-step tokens differ by a few hundred.
\emph{Salvage estimate}: for a truncated trajectory, we take the work it still needed to be the non-error tokens of a completed trajectory on the same question (median over completed siblings) minus the non-error tokens it had already produced; removing subtype $r$ would have let it finish if the tokens on its $r$-steps cover that shortfall. Of 441 truncated trajectories with a completed sibling, only 20 have a positive shortfall; per-subtype values are in \texttt{salvage\_estimate.csv}.
\begin{table*}[t]
\centering
\scriptsize
\setlength{\tabcolsep}{3pt}
\begin{adjustbox}{max width=\linewidth}
\begin{tabular}{@{}lrrrrrrrrr@{}}
\toprule
& \multicolumn{2}{c}{\textbf{Off-path \%}} & \multicolumn{2}{c}{\textbf{Dead-end \%}} & \multicolumn{2}{c}{\textbf{Failure-step \%}} & \multicolumn{3}{c}{\textbf{Salvage}} \\
\cmidrule(lr){2-3}\cmidrule(lr){4-5}\cmidrule(lr){6-7}\cmidrule(lr){8-10}
\textbf{Task} & corr. & incorr. & compl. & trunc. & compl. & trunc. & $n$ & rep. & all \\
\midrule
ZebraLogic & 38 & 52 & 25 & 58 & 5.5 & 6.7 & 179 & 3.4 & 3.4 \\
24~Game & 64 & 82 & 32 & 46 & 11.7 & 12.0 & 232 & 1.3 & 1.3 \\
PlanBench & 86 & 93 & 50 & 65 & 3.4 & 3.4 & 30 & 0.0 & 0.0 \\
Pooled & 62 & 91 & 43 & 56 & 5.4 & 7.5 & 441 & 2.0 & 2.0 \\
\bottomrule
\end{tabular}
\end{adjustbox}
\caption{Where the budget goes (Qwen3-8B; definitions in the text). Off-path: share of a trajectory's tokens on steps whose node is not on the committed answer's dependency path, for correct and incorrect trajectories. Dead-end: share of tokens on steps whose node the oracle marks unable to reach a correct solution and that carry no failure row, for completed and truncated trajectories. Failure-step: share of tokens on steps carrying any failure row. Salvage: of the $n$ truncated trajectories with a completed sibling on the same question, the share that would have finished had the tokens of its repetition steps (rep.) or of all its failure steps (all) been available.}
\label{tab:app-budget}
\end{table*}

Table~\ref{tab:app-budget} gives the three estimates per task.
\paragraph{Returns to dead regions.}
A dead region is a maximal subtree whose root the viability oracle marks unable to reach a correct solution, and a return is a later visit to the region after at least five steps on nodes outside it (Table~\ref{tab:app-dead-returns}).
Trajectories do come back to dead regions, most non-correct ones several times, but the returns hold about a sixth of the dead-end tokens; most returns enter a region the model never closed, and most returning first steps carry no failure row, so a return is mostly fresh work in the region rather than a repetition.
At a gap of one step, PlanBench's returns are one-step hops to the parent and back along the plan chain, which is why we read the five-step gap.
\begin{table}[t]
\centering
\footnotesize
\setlength{\tabcolsep}{3pt}
\begin{adjustbox}{max width=\linewidth}
\begin{tabular}{@{}l rrr r@{}}
\toprule
& ZebraLogic & 24 Game & PlanBench & Pooled \\
\midrule
\multicolumn{5}{@{}l}{\textit{Per dead region (\%)}} \\
re-entered after being left & 18 & 17 & 34 & 23 \\
dead-end tokens held by returns & 16 & 17 & 16 & 15.7 \\
returns into a region the model had closed & 33 & 49 & 13 & 22 \\
return's first step carries a repetition row & 30 & 44 & 2 & 13 \\
\midrule
\multicolumn{5}{@{}l}{\textit{Per trajectory, correct / incorrect / truncated}} \\
share with at least one return (\%) & 33 / 60 / 85 & 71 / 97 / 98 & 81 / 93 / 87 & -- \\
median number of returns & 0 / 1 / 2 & 2 / 10 / 14 & 4 / 10 / 14.5 & -- \\
median share of dead-end tokens on returns (\%) & 0 / 1 / 5 & 8 / 17 / 18 & 12 / 13 / 10 & -- \\
\bottomrule
\end{tabular}
\end{adjustbox}
\caption{\textbf{Returns to dead regions} (Qwen3-8B, 161{,}897 dead regions, gap of five steps). Pooled, 78\% of returns enter a region the model never closed and 87\% of returning first steps carry no failure row; the median gap before a return is 28 steps, and returns hold 9.3\% of dead-end tokens at a gap of twenty steps.}
\label{tab:app-dead-returns}
\end{table}

\subsection{Robustness to the engine-development problems}
\label{sec:app-engine-dev}
Every number in this part excludes the 40 problems per task on which we developed the parser and engine; recomputed with them (12{,}078 traces), every headline number moves in the second decimal or less, with one change of significance among reported subtypes (pooled Branch Abandonment under Model C, 0.83 [0.69, 1.02] without and 0.80 [0.66, 0.95] with them). The PlanBench schema validator records 618 node-level invariant violations among the 593{,}057 nodes of the main-text set (about 1 in 1{,}000): 344 parent--child kind pairs outside the schema and 274 nodes whose state differs from what the parent state and the step's effects imply. We record them without correcting them, and the nodes stay in every count.

\FloatBarrier
\section{Precursors: additional results}
\label{sec:app-precursors}
\begin{table}[t]
\centering
\scriptsize
\setlength{\tabcolsep}{3pt}
\begin{tabular}{@{}lrrrr@{}}
\toprule
\textbf{Subtype} & \textbf{State} & \textbf{Labels} & \textbf{Pos.} & \textbf{Prev.\ op} \\
\midrule
\multicolumn{5}{@{}l}{\emph{24~Game}} \\
\quad R1 Re-Derivation & \textbf{0.73} & 0.70 & 0.50 & 0.69 \\
\quad R2 Derive-after-Prune$^{\dagger}$ & 0.59 & 0.51 & 0.50 & 0.57 \\
\quad R3 Redundant Re-Verification & \textbf{0.83} & 0.81 & 0.58 & 0.77 \\
\quad R4 Subgoal Churn & \textbf{0.77} & 0.64 & 0.51 & 0.60 \\
\quad R5 Branch Abandonment & \textbf{0.91} & 0.81 & 0.56 & 0.63 \\
\quad R7 Post-Answer Attempt & \textbf{0.87} & 0.87 & 0.60 & 0.77 \\
\quad R9 Infeasible Derivation & \textbf{0.78} & 0.71 & 0.53 & 0.62 \\
\multicolumn{5}{@{}l}{\emph{ZebraLogic}} \\
\quad R1 Re-Derivation & \textbf{0.64} & 0.50 & 0.51 & 0.55 \\
\quad R2 Derive-after-Prune & \textbf{0.70} & 0.68 & 0.49 & 0.59 \\
\quad R3 Redundant Re-Verification$^{\dagger}$ & 0.70 & 0.66 & 0.56 & 0.70 \\
\quad R9 Infeasible Derivation$^{\dagger}$ & 0.64 & 0.64 & 0.54 & 0.61 \\
\multicolumn{5}{@{}l}{\emph{PlanBench}} \\
\quad R3 Redundant Re-Verification & \textbf{0.66} & 0.56 & 0.55 & 0.57 \\
\quad R9 Infeasible Derivation & \textbf{0.76} & -- & 0.55 & 0.61 \\
\quad R10 False Rejection$^{\dagger}$ & 0.62 & 0.57 & 0.57 & 0.59 \\
\quad R11 False Verdict & \textbf{0.62} & -- & 0.54 & 0.56 \\
\bottomrule
\end{tabular}
\caption{\textbf{The state before a step separates a failure from the steps where it could have happened.} Held-out AUROC of a classifier on the tree state over the preceding steps (State) against each subtype's matched opportunity steps, and the same classifier restricted to operation labels (Labels), to position in the trajectory (Pos.), and to the previous operation (Prev.\ op). Bold: state above the previous-operation baseline by more than 0.04; $^{\dagger}$: within 0.04, inconclusive. --: no label-only feature survived screening.}
\label{tab:precursors}
\end{table}

\begin{table}[t]
\centering
\scriptsize
\setlength{\tabcolsep}{3pt}
\begin{tabular}{@{}lrrrr@{}}
\toprule
\textbf{Subtype} & \textbf{State} & \textbf{Labels} & \textbf{Pos.} & \textbf{Prev.\ op} \\
\midrule
\multicolumn{5}{@{}l}{\emph{Pooled}} \\
\quad R1 Re-Derivation$^{\dagger}$ & 0.57 & 0.50 & 0.50 & 0.60 \\
\quad R2 Derive-after-Prune & \textbf{0.71} & 0.51 & 0.50 & 0.56 \\
\quad R3 Redundant Re-Verification$^{\dagger}$ & 0.74 & 0.73 & 0.58 & 0.70 \\
\quad R9 Infeasible Derivation & \textbf{0.73} & 0.53 & 0.53 & 0.54 \\
\bottomrule
\end{tabular}
\caption{Pooled cells (the three tasks together, with task indicators as features), as in Table~\ref{tab:precursors}.}
\label{tab:app-precursors-pooled}
\end{table}

\begin{figure}[t]
\centering
\caseblock[\linewidth]{Infeasible Derivation (R9), 24~Game, tiles 4, 6, 8, 10}{at most one constraint (remaining pool) declared on the branch over the last steps; no check for a long run}{
\cl{123}{Let me think again.}{constraints on branch: 1; last check 45 steps back}
\cl{124}{Alternatively, 6 / ((10 - 8)) = 6 / 2 = 3.}{uses 6, 10, 8; pool not restated}
\ev{125}{Then 8 * 3 = 24.}{constraints on branch: 1; reuses 8}
\cl{126}{Wait, but that would use 6, 10, 8, and then 4 is left.}{check: counts 8 once}
\cl{127}{So that's not using all four numbers.}{check fails on the unused 4, not on the reused 8}
}
\caption{\textbf{The state before an Infeasible Derivation.} Trace steps before the failure (step numbers from the trace); under the title, the state that predicts the failure on held-out problems; gray: that state's reading on each step; shaded: the failing step. Excerpts for the other subtypes are in Figure~\ref{fig:precursor-excerpts}.}
\label{fig:precursor-case-main}
\end{figure}

\begin{figure*}[t]
\centering

\caseblock{Branch Abandonment (R5), 24~Game, tiles 1, 7, 8, 12}{the leaving step is a derivation, not a check or a remark, from a frontier leaf with nothing tried under it and no verdict}{
\cl{11}{If I can get 12 multiplied by 2, but how to get 2 from 1 and 7?}{branch $12\times2$, depth 1}
\cl{12}{Well, 7 - 1 is 6, not 2.}{one option closed; branch open}
\cl{13}{7 divided by something?}{new node, depth 3}
\cl{14}{Maybe 7 - (something) equals 2?}{new node, depth 3}
\cl{15}{Let me think.}{no new node (1 step), no verdict}
\cl{16}{Alternatively, maybe using 8 and 12 together.}{no new node (2 steps), no verdict}
\ev{17}{For example, 12 - 8 = 4.}{a derivation, from the root: leaves the frontier leaf of step 14 unjudged; $12\times2$ was solvable ($12\times(8-7+1)$)}
}\\[6pt]
\caseblock{Post-Answer Attempt (R7), 24~Game, tiles 4, 6, 10, 11}{many steps since the branch's last check of the standing solution, no check in the last 5 steps}{
\cl{58}{So the answer would be (4 * 6) * (11 - 10).}{correct solution; last check 1 step back}
\cl{59}{But maybe the parentheses can be arranged differently.}{last check 2 steps back}
\cl{61}{Either way, the expression is valid.}{4 steps back}
\cl{62}{Let me make sure there are no other operations that might be needed.}{5 steps back}
\cl{63}{For example, maybe a division or subtraction that I haven't considered.}{6 steps back}
\cl{65}{Suppose I try other combinations.}{8 steps back; 0 checks in the last 5}
\ev{66}{For example, 11 + 10 = 21.}{new branch instead of a commit}
}\hfill
\caseblock{Redundant Re-Verification (R3), 24~Game, tiles 1, 1, 1, 8}{the node was already checked several times, the previous steps are checks of it}{
\cl{20}{Each tile is used once: yes, 8 once, each 1 once.}{check 1 of $8\times(1{+}1{+}1)$: passes}
\cl{22}{Let me check the calculation: 1+1+1 is 3, multiplied by 8 gives 24.}{check 3: passes}
\cl{23}{That seems correct.}{check 4: passes}
\cl{26}{So that's valid.}{tiles checked again: passes}
\cl{27}{Wait, but maybe there's a mistake here.}{no new information}
\cl{28}{Let me check again.}{}
\ev{30}{So the expression 8*(1+1+1) uses each of the four numbers once.}{check 6 of the same node, same content as check 1: passes}
}
\caption{\textbf{The state before a failure.} Four excerpts (step numbers from the trace). Under each title, the state that predicts the failure on held-out problems; gray: that state's reading on the step; shaded: the failure step.}
\label{fig:precursor-excerpts}
\end{figure*}

\begin{table}[t]
\centering
\scriptsize
\setlength{\tabcolsep}{3pt}
\begin{tabular}{@{}llrrrr@{}}
\toprule
\textbf{Pair} & \textbf{Task} & \textbf{Jaccard} & \textbf{State} & \textbf{Pos.} & \textbf{Prev.\ op} \\
\midrule
R1 vs R2 & ZebraLogic & 0.00 & 0.63 & 0.54 & 0.57 \\
R1 vs R2 & 24~Game & 0.00 & 0.65 & 0.54 & 0.62 \\
R1 vs R2 & Pooled & 0.14 & 0.83 & 0.60 & 0.60 \\
R1 vs R3 & ZebraLogic & 0.00 & 0.84 & 0.74 & 0.69 \\
R1 vs R3 & 24~Game & 0.00 & 0.88 & 0.78 & 0.77 \\
R1 vs R3 & Pooled & 0.13 & 0.81 & 0.81 & 0.72 \\
R5 vs R7 & 24~Game & 0.00 & 0.86 & 0.70 & 0.69 \\
R9 vs R10 & PlanBench & 0.00 & 0.71 & 0.58 & 0.64 \\
\bottomrule
\end{tabular}
\caption{Sibling subtypes arise after different states. Jaccard: overlap of the two subtypes' stable precursor sets. State: AUROC of a classifier on the union of the two sets at telling one sibling's occurrences from the other's, with the two baselines.}
\label{tab:app-precursors-siblings}
\end{table}

\begin{figure*}[t]
\centering
\caseblock{Re-Derivation (R1), 24~Game, tiles 1, 2, 7, 10}{no derivation and no new node on the branch in the last 3 steps}{
\cl{8}{10 and 2 could be multiplied to get 20, but then I need 4 more.}{derives $10\times2=20$}
\el
\cl{118}{Hmm, maybe (7 - something)/something.}{new node}
\cl{119}{Let me think.}{no new node (1 step)}
\cl{120}{Not sure.}{no new node (2 steps)}
\cl{121}{Wait, maybe using subtraction.}{no new node (3 steps)}
\cl{122}{Let me think.}{no new node (4 steps), no derivation}
\ev{123}{For example, 10 × 2 = 20.}{re-derives step 8, on a branch left 115 steps earlier}
}\hfill
\caseblock{Subgoal Churn (R4), 24~Game, tiles 8, 8, 8, 11}{switches in the last 5 steps with no prune on the branch}{
\cl{12}{For example, 8 * 3 = 24, but how do I get 3 from the remaining numbers?}{posts goal: 3 from 8, 8, 11}
\el
\cl{71}{Let's think of combining the numbers in a different way.}{switch 1, no prune}
\cl{72}{For example, (11 + 8) = 19.}{new branch}
\cl{73}{Then maybe combine with the other 8s?}{no new node}
\cl{74}{Not sure.}{no new node}
\cl{75}{Let me try.}{no new node; no prune since the switch}
\ev{76}{Another idea: Let's think of 24 as 8 * 3.}{switch 2 within 5 steps, no prune: re-posts the goal of step 12, options still open}
}\\[6pt]
\caseblock{Infeasible Derivation (R9), PlanBench Blocksworld}{every step on the branch applicable over the last 20 steps; the step then needs a fact its own earlier step made false}{
\cl{433}{Then, step 4: put down orange on the table.}{applicable; 23 valid steps in a row}
\cl{435}{Then, step 5: pick up red.}{applicable (25 in a row)}
\cl{436}{Step 6: stack red on top of orange (which is on the table).}{applicable (26); orange is no longer clear}
\cl{437}{Now, red is on top of orange.}{27}
\el
\cl{441}{So step 7: pick up yellow.}{applicable (31)}
\cl{442}{Step 8: stack yellow on blue.}{applicable (32)}
\cl{444}{Now, yellow is on top of blue.}{34 valid steps in a row}
\ev{445}{Then, step 9: pick up orange (which is on the table).}{inapplicable: red sits on orange since plan step 6 (step 436)}
\cl{446}{Step 10: stack orange on top of yellow.}{builds on it}
\cl{450}{Which is the goal.}{check passes; step 6 not re-read}
}\hfill
\caseblock{Infeasible Derivation (R9), ZebraLogic}{a case opened on the previous step (ASSUME); the derivation is the first step under it}{
\cl{778}{Case3: Carol is in house4.}{case opened}
\cl{783}{Therefore, Bob must be in 5.}{}
\cl{784}{But clue5 says Bob is not in house5.}{fixes: Bob is not in house 5}
\cl{786}{So Carol can't be in house4.}{case closed}
\cl{787}{Case4: Carol is in house6.}{ASSUME on the previous step: new case}
\ev{788}{Then Bob must be adjacent, so in house5.}{first step under the case; puts Bob in house 5 against clue 5 (step 784)}
\cl{789}{So Carol is in 6, Bob in 5.}{used}
\cl{792}{Clue13 says mystery is not in 5, which is okay here (Carol is in 6).}{check of clue 13 passes; clue 5 not re-read}
\cl{793}{Now, Bob is in house5.}{kept; never caught before the trace ends (step 803)}
}\\[6pt]
\caseblock{Derive-after-Prune (R2), ZebraLogic}{no prune in the last 10 steps; the derivation lands in a case ruled out earlier that no step re-opened}{
\cl{286}{But house 1's flower is lilies, so if the cruise lover is in house 1, then their flower would be carnations, which contradicts house 1's flower being lilies.}{prune: the cruise lover is not in house 1}
\el
\cl{298}{Also, house 1's flower is lilies.}{restated in the new case; last prune 8 steps back}
\cl{310}{Also, clue 13: tennis lover is cruise lover, and clue 3: cruise lover has carnations.}{20 steps without a prune}
\cl{316}{house 1: lilies}{26 steps without a prune}
\cl{319}{So, one of these houses has carnations (cruise lover).}{new node}
\cl{324}{So, the remaining vacations are cruise and beach.}{new node; 34 steps without a prune}
\ev{325}{So the cruise lover is in either house 1, 2, 3, but house 1's vacation is not determined yet.}{re-opens house 1 for the cruise lover, closed at step 286; no step re-opened it}
\cl{326}{But house 1's vacation could be cruise or beach.}{derives inside the closed case}
\cl{334}{Since Eric's vacation is city, and house 4 is mountain, the remaining are cruise and beach for houses 1, 2, 3.}{still open 9 steps later; closed again only at step 340}
}
\caption{Further excerpts, as in Figure~\ref{fig:precursor-excerpts}.}
\label{fig:app-precursor-cases}
\end{figure*}

\begin{table*}[t]
\centering
\footnotesize
\setlength{\tabcolsep}{4pt}
\begin{tabularx}{\textwidth}{@{}l >{\raggedright\arraybackslash}p{4.6cm} X@{}}
\toprule
\textbf{Feature} & \textbf{Condition the stand-in reads} & \textbf{Mitigation it gates, and what the gate changes} \\
\midrule
Branch content & the constraints fixed on the branch within the last ten steps & a thinking intervention \citep{wu2025thinkingintervention}, a chain-of-verification \citep{dhuliawala2023chainofverification} or a process reward \citep{lightman2024verify,wang2024mathshepherd}: the inserted text or the verification question names those constraints, so the check lands on the constraint an Infeasible Derivation violates \\
Frontier status & whether the current frontier leaf has a verdict & the thought-switch penalty \citep{wang2025underthinking}: applied only while the frontier leaf has no verdict, so it stops Branch Abandonment and spares the switches that leave a ruled-out branch, which an unconditional penalty suppresses as well \\
Frontier status & whether a checked complete solution stands & budget forcing \citep{muennighoff2025s1} and confidence-gated early exit \citep{yang2025deer,zhang2025probing}: generation stops when a checked complete solution stands rather than at a length or confidence threshold, which ends the Post-Answer Attempt without cutting off unfinished search \\
Recent operations & the branch's last operation and its pruned status & steering vectors for verification and backtracking \citep{venhoff2025steering,ward2025repurposes}: the verification vector is damped right after a check, which removes Redundant Re-Verification, and the backtracking vector is damped on a pruned branch, which removes Derive-after-Prune \\
\midrule
Offline & the return rows and the dead-end subtrees opened at a departure & spans to remove when compressing trajectories for fine-tuning \citep{xia2025tokenskip,zhao2025pruneonlogic,munkhbat2025selftraining}, replacing token-importance or length criteria \\
Offline & the per-step failure rows & a structure-aware penalty for length-harmonizing reinforcement learning \citep{luo2025o1pruner,chen2025overthinking} that charges a repeated operation rather than every token past the answer \\
Offline & the per-trajectory failure counts & selection among sampled trajectories \citep{hassid2025dontoverthink} by what they contain rather than by length \\
\bottomrule
\end{tabularx}
\caption{\textbf{The two routes, feature by feature.} Route 1 (first four rows): a stand-in trained on \arbor{}'s labels reads the condition that precedes the failure and gates an existing inference-time mitigation on it. Route 2 (last three rows): \arbor{}'s labels, computed offline, replace length and token criteria in training-side methods.}
\label{tab:app-routes}
\end{table*}

\paragraph{Opportunity sets and matching.}
An opportunity step is a step on which the subtype's precondition holds on the pre-step state (containment of the occurrences is 100\% on every analyzed row). We match five controls per occurrence on task, parser label and trajectory tenth, preferring the same trajectory, then the same problem. The features are 639 quantities of the pre-step state and the operation labels over windows of 1 to 20 steps and since the branch start. We screen them by standardized mean difference with Benjamini--Hochberg $q=0.1$, keep the top 60, and fit gradient-boosted trees (LightGBM; at most 300 trees with early stopping on grouped five-fold cross-validation over the development problems, learning rate 0.05, 15 leaves, minimum 20 samples per leaf, 0.8 row and column subsampling, L2 regularization 1.0, class-balanced) with stability selection; cells with 100 to 300 development events use nested cross-validation. We hold out 40 development problems per task and report every AUROC on the remaining problems. Two LLM judges, with a third as tie-break, judged 20 sampled opportunity cards per subtype against the trace text; acceptance 1.00 (Post-Answer Attempt), 0.85 (Derive-after-Prune, Redundant Re-Verification, Subgoal Churn, False Rejection), 0.75 (Branch Abandonment, Infeasible Derivation), 0.70 (False Verdict), 0.60 (Re-Derivation). 13 of 19 cells use a tightened opportunity set (we add a recency condition so that the detector's own firing quantity differs between occurrences and controls); the six cells on the main set are Re-Derivation on the 24~Game and pooled, Redundant Re-Verification and Branch Abandonment on the 24~Game, Derive-after-Prune on ZebraLogic, and Infeasible Derivation on PlanBench.
\paragraph{Rare subtypes.}
Seven of the eleven subtypes have enough occurrences for a held-out estimate, the four repetition subtypes, Branch Abandonment, Post-Answer Attempt and Infeasible Derivation (the rare ones in Appendix~\ref{sec:app-precursors}); we model False Rejection on PlanBench, but the estimate is inconclusive, and False Verdict there separates only weakly.
We observe every subtype in the taxonomy at least 76 times at 8B. Five task-rows fall below 10\% prevalence (ZebraLogic Post-Answer Attempt 8.4\% and Premature Finalization 3.2\%; PlanBench Subgoal Abandonment 4.5\%, Post-Answer Attempt 7.5\% and Premature Finalization 1.6\%); all five are present at 32B and, except PlanBench Premature Finalization, at 1.7B. Three are rare because the model rarely takes a common opportunity (ZebraLogic Post-Answer Attempt: the opportunity holds in 72\% of trajectories and is taken at 0.0022 per step; PlanBench Subgoal Abandonment 99.8\% and 0.0003; PlanBench Post-Answer Attempt 34\% and 0.0023), one because a commit run before the answer is itself uncommon and then rarely invalid (ZebraLogic Premature Finalization: 69\% of trajectories, 0.030), and PlanBench Premature Finalization is the inverse on a lower-bound count (an in-reasoning commit run in 1.9\% of trajectories, invalid in 75\% of them).
\begin{table*}[t]
\centering
\scriptsize
\setlength{\tabcolsep}{3pt}
\begin{adjustbox}{max width=\linewidth}
\begin{tabular}{@{}llp{0.5cm}p{7.2cm}rr@{}}
\toprule
\textbf{Subtype} & \textbf{Task} & \textbf{Set} & \textbf{Stable precursor, in words} & \textbf{Prec.} & \textbf{Rec.} \\
\midrule
R1 Re-Derivation & 24~Game & main & no derivation and no new node on the branch in the last 3 steps & 0.32 & 0.58 \\
R1 Re-Derivation & ZebraLogic & tight & a node was created on the branch within the last 10 steps & 0.33 & 0.20 \\
R1 Re-Derivation & Pooled & main & open subgoals falling while the branch adds no node & 0.17 & 0.66 \\
R2 Derive-after-Prune & 24~Game & tight & a long run since the branch's last check, confirm or assumption; no node created on the previous step & 0.20 & 0.70 \\
R2 Derive-after-Prune & ZebraLogic & main & no prune in the last 10 steps & 1.00 & 0.31 \\
R2 Derive-after-Prune & Pooled & tight & a long run since the trajectory's last prune relative to recent node creation & 0.16 & 0.59 \\
R3 Redundant Re-Verification & 24~Game & main & the branch's last check is recent and the previous step is a check & 0.83 & 0.19 \\
R3 Redundant Re-Verification & ZebraLogic & tight & many remarks (META) in the last 20 steps and a long run since the branch's last check & 0.38 & 0.05 \\
R3 Redundant Re-Verification & PlanBench & tight & a check was recent on the branch while few nodes were created in the last 20 steps & 0.37 & 0.10 \\
R3 Redundant Re-Verification & Pooled & tight & the branch's last check is recent relative to the trajectory's & 0.49 & 0.36 \\
R4 Subgoal Churn & 24~Game & tight & switches in the last 5 steps without a prune on the branch & 0.49 & 0.59 \\
R5 Branch Abandonment & 24~Game & main & the step is a derivation (not a check, restatement or remark) at recent depth above 1.8 & 0.47 & 0.64 \\
R7 Post-Answer Attempt & 24~Game & tight & a long run since the branch's last check and no check in the last 3 to 5 steps & 0.30 & 0.42 \\
R9 Infeasible Derivation & 24~Game & tight & at most one constraint (remaining pool) declared on the branch over the last 3 to 20 steps & 0.58 & 0.51 \\
R9 Infeasible Derivation & ZebraLogic & tight & an assumption (new case) on the previous step; the derivation is the first step under it & 0.50 & 0.17 \\
R9 Infeasible Derivation & PlanBench & main & the branch has been oracle-viable over the last 20 steps & 0.35 & 0.84 \\
R9 Infeasible Derivation & Pooled & tight & the branch has been viable recently and its last check is recent & 0.35 & 0.42 \\
R10 False Rejection & PlanBench & tight & shorter steps than average over the last 10 to 20 steps & 0.26 & 0.80 \\
R11 False Verdict & PlanBench & tight & high recent fan-out on the branch & 0.52 & 0.12 \\
\bottomrule
\end{tabular}
\end{adjustbox}
\caption{The stable precursor of each modeled cell in words (the feature set selected in every refit), the opportunity set it was fit on (main, or tightened by a recency condition), and the test precision and recall of the best single rule on that set.}
\label{tab:app-precursor-rules}
\end{table*}

\begin{table*}[t]
\centering
\footnotesize
\setlength{\tabcolsep}{3pt}
\begin{adjustbox}{max width=\textwidth}
\begin{tabular}{@{}l rrrr rrrr rrrr@{}}
\toprule
& \multicolumn{4}{c}{ZebraLogic} & \multicolumn{4}{c}{24 Game} & \multicolumn{4}{c}{PlanBench} \\
\cmidrule(lr){2-5} \cmidrule(lr){6-9} \cmidrule(lr){10-13}
Subtype & visits & age & path \% & sub. & visits & age & path \% & sub. & visits & age & path \% & sub. \\
\midrule
R1 Re-Derivation & 2 / 2 & 6 / 2 & 21 / 28 & 9 / 17 & 2 / 0 & 90 / 0 & 4 / 7 & 6 / 2 & 3 / 2 & 99 / 3 & 15 / 12 & 23 / 5 \\
R2 Derive-after-Prune & 3 / 2 & 23 / 3 & 4 / 24 & 2 / 14 & 2 / 0 & 77 / 0 & 1 / 7 & 1 / 2 & 0 / 0 & 0 / 0 & 15 / 10 & 8 / 5 \\
R3 Redundant Re-Verification & 18 / 12 & 46 / 17 & 46 / 43 & 1 / 2 & 38 / 3 & 54 / 5 & 67 / 29 & 1 / 1 & 7 / 5 & 25 / 6 & 29 / 15 & 5 / 4 \\
R4 Subgoal Churn & -- & -- & -- & -- & 2 / 0 & 96 / 0 & 3 / 7 & 7 / 2 & 4 / 1 & 39 / 1 & 4 / 9 & 3 / 4 \\
R5 Branch Abandonment & 4 / 7 & 9 / 8 & 23 / 57 & 2 / 12 & 2 / 11 & 5 / 17 & 4 / 53 & 2 / 1 & 3 / 4 & 8 / 5 & 7 / 14 & 3 / 6 \\
R6 Subgoal Abandonment & -- & -- & -- & -- & 2 / 9 & 7 / 13 & 0 / 45 & 2 / 1 & -- & -- & -- & -- \\
R7 Post-Answer Attempt & 29 / 30 & 30 / 41 & 44 / 73 & 1 / 1 & 30 / 13 & 64 / 19 & 67 / 48 & 2 / 1 & 7 / 3 & 94 / 6 & 30 / 23 & 3 / 4 \\
R8 Premature Finalization & 33 / 27 & 43 / 34 & 96 / 96 & 1 / 1 & 8 / 39 & 12 / 52 & 51 / 75 & 1 / 1 & 0 / 13 & 0 / 187 & 75 / 77 & 2 / 26 \\
R9 Infeasible Derivation & 0 / 0 & 0 / 0 & 6 / 32 & 4 / 20 & 0 / 0 & 0 / 0 & 15 / 12 & 1 / 2 & 0 / 0 & 0 / 0 & 9 / 10 & 7 / 5 \\
R10 False Rejection & 2 / 2 & 19 / 6 & 11 / 5 & 6 / 2 & 2 / 1 & 3 / 1 & 5 / 0 & 1 / 1 & 2 / 2 & 4 / 4 & 1 / 1 & 1 / 1 \\
R11 False Verdict & 14 / 21 & 16 / 23 & 37 / 63 & 3 / 2 & -- & -- & -- & -- & 2 / 3 & 4 / 5 & 11 / 14 & 3 / 3 \\
\bottomrule
\end{tabular}
\end{adjustbox}
\caption{\textbf{The node at the event, against the nodes at which the row could have fired.} Each cell is event median / opportunity-step median (Qwen3-8B): earlier visits to the node, node age in steps, share of events whose node lies on the committed answer's path, and final subtree size. This is the univariate view of the contrast that \S\ref{sec:results-precursors} models; the full feature set and held-out AUROC are in Table~\ref{tab:precursors}. --: row undefined on the task. R8 on PlanBench has 76 events against 26 opportunity steps.}
\label{tab:app-prestate}
\end{table*}

Table~\ref{tab:app-prestate} gives the univariate form of the same contrast: four descriptive quantities of the node at the event against the row's opportunity steps, per task.
Table~\ref{tab:app-precursor-rules} gives the stable precursor of every modeled cell in words, with the test precision and recall of the best single rule; Table~\ref{tab:app-precursors-siblings} gives the sibling-pair separation behind the last sentence of the first paragraph of \S\ref{sec:results-precursors}, and Figures~\ref{fig:precursor-case-main}, \ref{fig:precursor-excerpts} and~\ref{fig:app-precursor-cases} the excerpts.
\paragraph{Pre-states by subtype.}
The pre-states behind the three points of \S\ref{sec:results-precursors} are these. An Infeasible Derivation violates a constraint the model itself fixed within the previous ten steps in about 3/4 of cases, on the 24~Game on a branch where it has stopped restating what remains, and the derivation is then left unverified in about half of cases and usually passed by the first verification that reaches it. A Branch Abandonment leaves on a derivation rather than a verification, from a frontier leaf with nothing tried under it and no verdict on it; a Post-Answer Attempt leaves a solution's branch after a run with no verification of the solution. A Redundant Re-Verification re-verifies a node already verified several times, shortly after another verification of it; a Re-Derivation comes when the branch has produced no derivation and no new node in its last three steps; a Derive-after-Prune comes inside a branch still closed, on the 24~Game after a long run since the branch's last verification or assumption and on ZebraLogic with no prune in the last ten steps; Subgoal Churn follows switching without pruning. These states tell apart siblings of one failure mode: a stalled branch before a Re-Derivation against a long run since the last prune before a Derive-after-Prune, and a young frontier leaf before a Branch Abandonment against a solution left unverified before a Post-Answer Attempt.
\paragraph{Routes, point by point.}
Table~\ref{tab:app-routes} lists, for each point, the condition a stand-in trained on \arbor{}'s labels reads and the existing mitigation it gates, and for the offline route the criterion \arbor{}'s labels replace.

\FloatBarrier

\end{document}